\documentclass[11pt]{article}

\usepackage[final]{acl}

\usepackage{times}
\usepackage{latexsym}

\usepackage[T1]{fontenc}

\usepackage[utf8]{inputenc}

\usepackage{microtype}

\usepackage{inconsolata}

\usepackage{graphicx}
\usepackage{amssymb}
\usepackage{amsmath}
\usepackage{algorithm}
\usepackage{algorithmic}
\usepackage{booktabs}
\usepackage{cleveref}
\usepackage{subcaption}
\usepackage{makecell}
\usepackage{multirow}
\usepackage{bm}

\usepackage{tcolorbox}

\newtcolorbox{conceptbox}{
  colback=gray!5,
  colframe=gray!50,
  boxrule=0.5pt,
  arc=3pt,
  left=4pt,
  right=4pt,
  top=4pt,
  bottom=4pt
}

\title{Forget What Matters, Keep the Rest: \\Selective Unlearning of Informative Tokens}

\author{
 Seunghee Koh\textsuperscript{1}\qquad 
 Sunghyun Baek\textsuperscript{1}\qquad 
 Youngdong Kim\textsuperscript{2}\thanks{Co-corresponding authors}\qquad 
 Junmo Kim\textsuperscript{1}\footnotemark[2]
\\
\\
 \textsuperscript{1}Korea Advanced Institute of Science and Technology, South Korea\\
 \textsuperscript{2}Hanbat National University, South Korea
\\
 {\tt\small \{seunghee1215, baeksh, junmo.kim\}@kaist.ac.kr} \qquad {\tt\small ydkim1293@hanbat.ac.kr}
}

\begin{document}
\maketitle
\begin{abstract}
Unlearning in large language models (LLMs) has emerged as a promising safeguard against adversarial behaviors. When the forgetting loss is applied uniformly without considering token-level semantic importance, model utility can be unnecessarily degraded. Recent studies have explored token-wise loss regularizers that prioritize informative tokens, but largely rely on ground-truth confidence or external linguistic parsers, which limits their ability to capture contextual information or the model’s overall predictive state.
Intuitively, function words like “the” primarily serve syntactic roles and are highly predictable with little ambiguity, but informative words admit multiple plausible alternatives with greater uncertainty.
Based on this intuition, we propose Entropy-guided Token Weighting (ETW), a token-level unlearning regularizer that uses entropy of the predictive distribution as a proxy for token informativeness. 
We demonstrate that informative tokens tend to have higher entropy, whereas structural tokens tend to have lower entropy. This behavior enables ETW to achieve more effective unlearning while better preserving model utility than existing token-level approaches.

\end{abstract}

\section{Introduction}
Machine unlearning \cite{machine_unlearning,eternal_sunshine,mixed_privacy,dtl} aims to train deep neural networks to selectively remove specific knowledge while retaining other knowledge. It has emerged as a promising approach for removing portions of training corpora in large language models (LLMs), where it can serve as a defense against membership inference, jailbreak, and red-teaming attacks \cite{minkpp,jailbreak,red_teaming}. In LLMs, unlearning follows naturally from the next-token prediction objective, which decomposes into per-token losses, motivating token-wise loss reweighting as a principled unlearning strategy. Prior work often uses ground-truth confidence as a token-level signal for unlearning \cite{wga,satimp}. Yet, confidence-based signals alone are insufficient to determine how strongly each token should be reweighted, leaving room for richer representational signals to guide token-wise penalties.

\begin{figure}
\centering
\includegraphics[width=\linewidth]{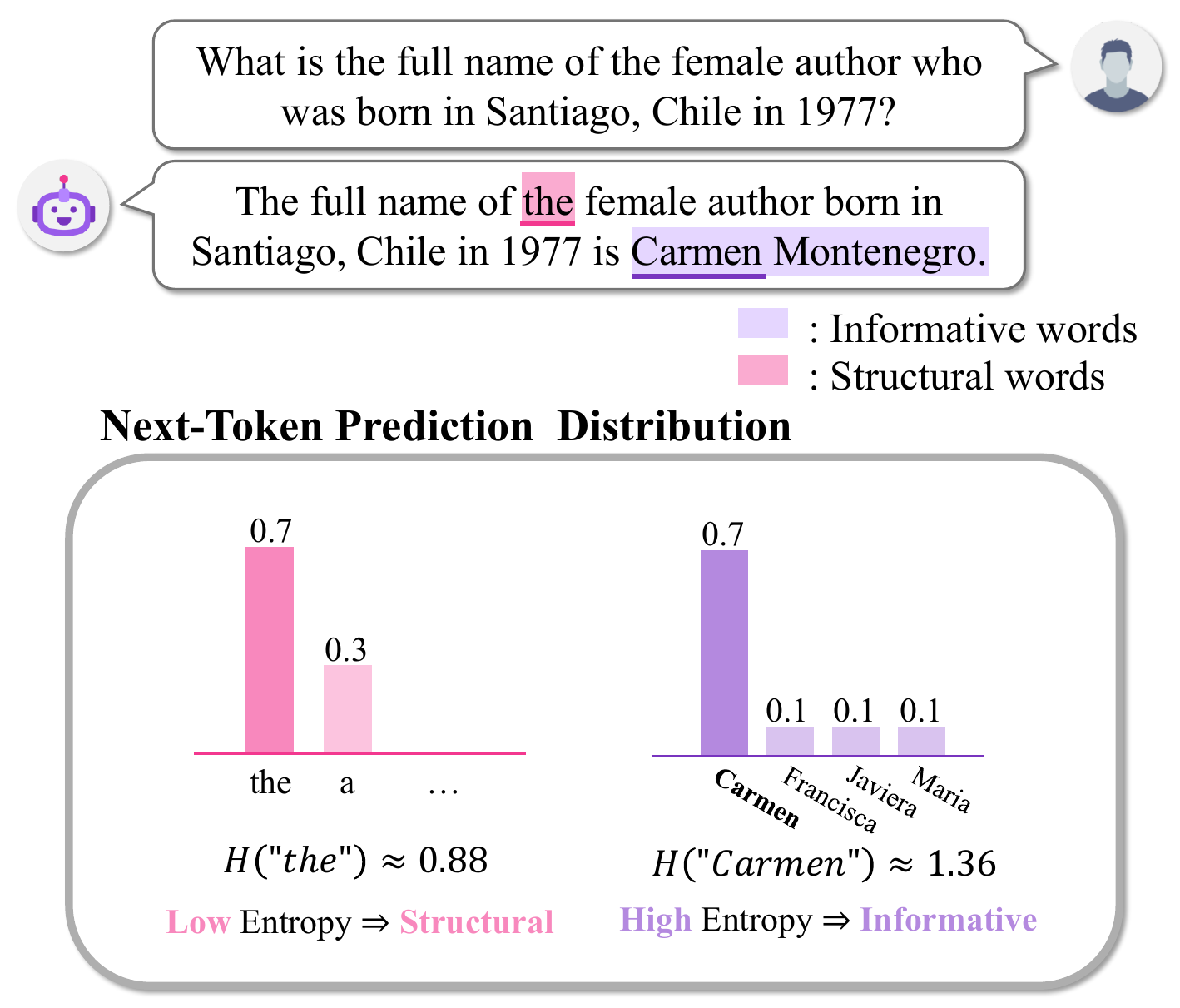}
\caption{ Motivation of Entropy-guided Token Weighting (ETW). Informative tokens such as “Carmen” exhibit higher entropy due to multiple plausible alternatives, while structural tokens such as “the” have lower entropy because they exhibit little ambiguity.}
\label{fig:example}
\end{figure}

A key question for effective unlearning is which tokens should be penalized more strongly while minimizing utility degradation. We distinguish between \emph{informative tokens}, which carry important semantic content, and \emph{structural tokens}, which serve mainly syntactic or repetitive functions. As illustrated in \Cref{fig:example}, informative tokens such as the proper noun “Carmen” admit multiple plausible alternatives and thus involve greater predictive uncertainty, whereas structural tokens such as the function word “the” leave little ambiguity in next-token prediction. This contrast suggests that entropy over the predictive distribution provides a useful signal for assessing token informativeness, even when ground-truth confidence is similar.

Based on this intuition, we propose \emph{Entropy-guided Token Weighting} (ETW), a token-wise regularizer that penalizes high-entropy tokens and downweights low-entropy ones. Unlike prior approaches that rely solely on the confidence of the ground-truth token, ETW quantifies informativeness by leveraging the model’s predicted distribution over all candidates in the vocabulary. Specifically, entropy serves as a richer proxy for informativeness by reflecting how probability mass is distributed among competing alternatives.

ETW more accurately identifies informative tokens than existing token-level regularizers. This improved identification enables effective forgetting while preserving model utility across a range of unlearning settings. The model unlearned with ETW also produces informative token prediction probabilities closest to those of the retrained model.

The main contributions of this paper are as follows:
\begin{itemize}
\item We introduce Entropy-guided Token Weighting (ETW), a token-wise regularizer for LLM unlearning that leverages entropy over the overall predictive distribution, rather than relying solely on ground-truth confidence.
\item We show that ETW is an effective measure for distinguishing informative tokens from structural tokens to identify informative tokens for selective penalization.
\item By focusing on informative tokens during unlearning, ETW consistently achieves effective forgetting while preserving model utility.
\end{itemize}

\section{Related Works}

\paragraph{Sample-Level LLM Unlearning}
LLM unlearning updates model parameters to selectively remove knowledge associated with a target corpus while aiming to preserve generalization. A basic approach applies gradient ascent (GA) loss \cite{tofu} to penalize the ground-truth token in the output space. Variants such as LOKU \cite{loku} and UNDIAL \cite{undial} redirect probability mass toward second-best tokens through loss or logit modulation.

DPO-inspired methods contrast target-corpus responses as losing answers against substitute responses as winning answers \cite{dpo,altpo}, which requires additional generation for winning answers. Although variants such as NPO \cite{npo} and SimNPO \cite{simnpo} remove explicit winning-answer dependencies, their objectives can ultimately be formulated as weighted GA. Except for SimNPO, they usually rely on a reference model.

RMU perturbs representations of the forget set, but is sensitive to perturbation design \cite{wga} and relies on reference-model distillation for retaining knowledge.

\paragraph{Token-Level LLM Unlearning}
Several works study token-level LLM unlearning through token-wise regularizers built on gradient ascent (GA)–based objectives, which modulate unlearning strength across tokens. WGA and TNPO \cite{wga} down-weight low-confidence tokens to control GA divergence, but penalize tokens solely based on confidence and fail to account for semantic importance within a forget sample.

SatImp \cite{satimp} and FUNDIAL \cite{undial} aim to reduce unnecessary degradation of general language capabilities by prioritizing informative tokens. SatImp reweights tokens using confidence and its complement, while FUNDIAL, a variant of UNDIAL, introduces a linguistically motivated strategy based on spaCy \cite{spacy}, hypothesizing that nouns and named entities encode core knowledge. However, SatImp relies on confidence values that are insufficient for identifying informative contents, and FUNDIAL depends on part-of-speech categories that do not explicitly incorporate contextual information.

ETW leverages entropy derived from the full candidate vocabulary distribution to selectively penalize informative tokens, rather than relying on confidence or coarse linguistic categories. 

\paragraph{Token-level Control}

Beyond unlearning, token-level control has been explored in LLMs across a range of tasks, including informative-token selection in supervised fine-tuning \cite{token_cleaning} and entropy-based token weighting in reinforcement learning with verifiable rewards (RLVR) \cite{beyond_80_20,ent_rl}.

For unlearning, we introduce an entropy-based token weighting scheme and demonstrate why entropy is particularly well-suited to this task (\Cref{sec:analysis}). Our method achieves effective forgetting while preserving model utility.
\section{Entropy-guided Token Weighting}
\subsection{Problem Setting}

We consider an autoregressive language model parameterized by $\bm{\theta}$ and trained on a sequence with a prompt $\mathbf{x} = (x_1, \ldots, x_m)$ with length $m$ and a completion $\mathbf{y} = (y_1, \ldots, y_n)$ with length $n$.

LLM unlearning aims to selectively forget targeted data while preserving knowledge from the retained data, which requires both a forgetting loss and a knowledge-retaining loss.

For {knowledge retaining}, we use the standard cross-entropy loss:
\begin{equation}
\mathcal{L}_{r}(\mathbf{y}|\mathbf{x}; \boldsymbol{\theta})
= - \log p\!\left(\mathbf{y}|\mathbf{x}; \boldsymbol{\theta}\right).
\end{equation}

For knowledge forgetting, we adopt the negative cross-entropy loss, also known as the gradient-ascent (GA) objective, whose token-wise decomposition is given by
\begin{equation}
\begin{aligned}
\mathcal{L}_{\mathrm{GA}}(\mathbf{y}|\mathbf{x}; \bm{\theta})
&= \log p(\mathbf{y}|\mathbf{x}; \bm{\theta}) \\
&= \sum_{i=1}^{n} \log p(y_i|y_{<i}, \mathbf{x} ; \bm{\theta}).
\end{aligned}
\end{equation}
Accordingly, we use the following token-wise weighted GA loss with weight $\omega_i(\mathbf{x}, \mathbf{y})$ as the primary objective:
\begin{equation}
\mathcal{L}_{f}(\mathbf{y}|\mathbf{x}; \bm{\theta})
= \sum_{i=1}^{n} \omega_i(\mathbf{x}, \mathbf{y}) \,
\log p(y_i|y_{<i},\mathbf{x}; \bm{\theta}).
\label{eq:weighted_ga}
\end{equation}

Given a retaining sample $(\mathbf{x}_r,\mathbf{y}_r) \in \mathcal{D}_r$ and a forgetting sample $(\mathbf{x}_f,\mathbf{y}_f) \in \mathcal{D}_f$, the overall unlearning objective is defined as
\begin{equation}
\mathcal{L}(\mathbf{y}|\mathbf{x}; \bm{\theta})
=
\mathcal{L}_r(\mathbf{y}_r|\mathbf{x}_r; \bm{\theta})
+
\lambda \mathcal{L}_f(\mathbf{y}_f|\mathbf{x}_f; \bm{\theta}),
\label{eq:unlearning_objective}
\end{equation}
where $\lambda$ controls the degree of knowledge removal associated with forgetting loss.

\begin{table*}[t]
\centering
\small
\setcellgapes{3pt}
\makegapedcells
\setlength{\tabcolsep}{6pt}
\renewcommand{\arraystretch}{0.5}
\begin{tabular}{l l l}
\toprule
Regularizer & \multicolumn{1}{c}{Explanation} & \multicolumn{1}{c}{Formulation} \\
\midrule
\multicolumn{3}{c}{\textit{Confidence-based token regularizer}} \\
\makecell[c]{WGA \\ \cite{wga}} &
\makecell[l]{Penalize high-confidence tokens \\ via exponentiated confidence} &
\(\omega^{\mathrm{WGA}}_i(\mathbf{x},\mathbf{y}; \bm{\theta)} = p(y_i|y_{<i},\mathbf{x}; \hat{\theta})^{\alpha}\) \\
\makecell[c]{Imp \\ \cite{satimp}}&
\makecell[l]{Penalize low-confidence tokens \\ using confidence complement} &
\(\omega^{\mathrm{Imp}}_i(\mathbf{x},\mathbf{y}; \bm{\theta}) = 1 - p(y_i|y_{<i},\mathbf{x}; \hat{\theta})\) \\
\makecell[c]{SatImp \\ \cite{satimp}}&
\makecell[l]{Token penalization via joint confidence \\ and confidence-complement weighting } &
\(\omega^{\mathrm{SatImp}}_i(\mathbf{x},\mathbf{y}; \bm{\theta}) =
p(y_i|y_{<i},\mathbf{x}; \hat{\bm\theta})^{\alpha}
\left(
1 - p(y_i|y_{<i},\mathbf{x}; \hat{\bm\theta})
\right)\) \\
\makecell[c]{TNPO \\ \cite{wga}}&
\makecell[l]{Penalize tokens with small confidence \\ deviation from a reference model} &
\(\omega^{\mathrm{TNPO}}_i(\mathbf{x},\mathbf{y}; \bm{\theta}) =
\dfrac{
2\, p(y_i|y_{<i},\mathbf{x}; \hat{\bm\theta})^{\beta}
}{
p(y_i|y_{<i},\mathbf{x}; \hat{\bm\theta})^{\beta}
+
p(y_i|y_{<i},\mathbf{x}; \bm\theta_{ref})^{\beta}
}\) \\
\midrule
\multicolumn{3}{c}{\textit{Linguistic-based token regularizer}} \\
\makecell[c]{SCN \\ \cite{undial}}&
\makecell[l]{Linguistic noun-based token selection \\ using an external parser (spaCy)} &
\(\omega^{\mathrm{SCN}}_i(\mathbf{x},\mathbf{y}) =
\mathbb{I}\!\left[
y_i \in \mathcal{E}_{\mathrm{Noun}}^{\mathrm{spaCy}}
\right],\ 
\mathcal{E}_{\mathrm{Noun}}^{\mathrm{spaCy}} \subseteq \mathcal{V}\) \\
\makecell[c]{SCE \\ \cite{undial}}&
\makecell[l]{Linguistic entity-based token selection \\ using an external parser (spaCy)} &
\(\omega^{\mathrm{SCE}}_i(\mathbf{x},\mathbf{y}) =
\mathbb{I}\!\left[
y_i \in \mathcal{E}_{\mathrm{Entity}}^{\mathrm{spaCy}}
\right],\ 
\mathcal{E}_{\mathrm{Entity}}^{\mathrm{spaCy}} \subseteq \mathcal{V}\) \\
\bottomrule
\end{tabular}
\caption{The formulation and explanation of existing token regularizers. $\bm\theta_{ref}$ denotes the reference model.}
\label{tab:token_regularizers}
\end{table*}

\subsection{The Formulation of Entropy-guided Token Weighting}
\label{sec:etw_formulation}
Within a single forget sample, tokens can be broadly categorized into two types:
\emph{informative tokens} and \emph{structural tokens}, defined as follows.

\begin{conceptbox}
\textbf{Informative Tokens} carry the core semantic content of the answer.

\textbf{Structural Tokens} primarily serve syntactic or repetitive roles,
such as function words or prompt mentions repeated in the completion.
\end{conceptbox}

As it is essential to selectively forget informative content while preserving overall utility, we aim to apply stronger penalties to informative tokens. We hypothesize that token-wise entropy can indicate token informativeness. Specifically, entropy measures how uncertain the model is about its next-token prediction. If the model is highly confident at a given position, the token is likely to be structural and carries limited semantic information. In contrast, if the model is uncertain and distributes probability mass across multiple candidate tokens, the position is more likely to encode informative or knowledge-related content.

Formally, the token-wise entropy for the $i$-th token in a completion $\mathbf{y}$ is defined over the vocabulary $\mathcal{V}$ as
\begin{align} 
&H(y_i|y_{<i},\mathbf{x};\bm{\theta}) \notag\\ 
&= - \sum_{v \in \mathcal{V}} p\!\left(v|y_{<i},\mathbf{x}; \bm{\theta}\right) \log p\!\left(v|y_{<i},\mathbf{x}; \bm{\theta}\right), 
\end{align}

From a mathematical perspective, even for a fixed ground-truth (GT) confidence $p_i := p(y_i|y_{<i},\mathbf{x};\bm{\theta})$, the range of entropy values can vary substantially depending on the distribution of non-GT tokens. The minimum achievable entropy occurs when the remaining probability mass $(1-p_i)$ is entirely concentrated on the second-best token:
\begin{equation}
H_{\min}(p_i)
= - p_i \log p_i - (1-p_i)\log(1-p_i).
\end{equation}
The maximum entropy is achieved when the residual probability mass $(1-p_i)$ is uniformly distributed across the remaining $|\mathcal{V}|-1$ vocabulary tokens:
\begin{equation}
H_{\max}(p_i)
= - p_i \log p_i - (1-p_i)\log\frac{1-p_i}{|\mathcal{V}|-1}.
\end{equation}

The fact that the distribution over non-ground-truth tokens governs the gap between $H_{\min}(p_i)$ and $H_{\max}(p_i)$ demonstrates that entropy offers a significantly richer representational range than measures based solely on ground-truth confidence.

To this end, we propose \emph{Entropy-guided Token Weighting} (ETW), a token-wise reweighting scheme derived from the model’s next-token predictive distribution. We normalize the entropy weights of completion tokens so that their sum equals the completion length $n$. This normalization preserves the overall scale of the unlearning loss by redistributing unlearning strength across tokens within a sample. Formally, ETW is defined as
\begin{equation}
{\omega}_i^{\mathrm{ETW}}(\mathbf{x},\mathbf{y}; \bm{\theta})
=
\frac{
n \cdot H(y_i|y_{<i},\mathbf{x};\hat{\bm{\theta}})
}{
\sum_{j=1}^{n} H(y_j|y_{<j},\mathbf{x};\hat{\bm{\theta}})
},
\end{equation}
where $\hat{\bm{\theta}}$ is a stop-gradient copy of the model parameters $\bm{\theta}$, such that
$\hat{\bm{\theta}}$ takes the value of $\bm{\theta}$ but $\frac{\partial \hat{\bm{\theta}}}{\partial \bm{\theta}} = 0$ during backpropagation.

Using $\hat{\bm{\theta}}$ ensures that the ETW computation does not affect gradient updates. Note that the probabilities used to compute entropy are obtained by applying softmax to the logits with temperature $T$. The resulting weights $\omega_i^{\mathrm{ETW}}$ are then applied to the token-wise forgetting loss in \Cref{eq:weighted_ga}.

\begin{figure}[t]
\centering
        \includegraphics[width=\linewidth]{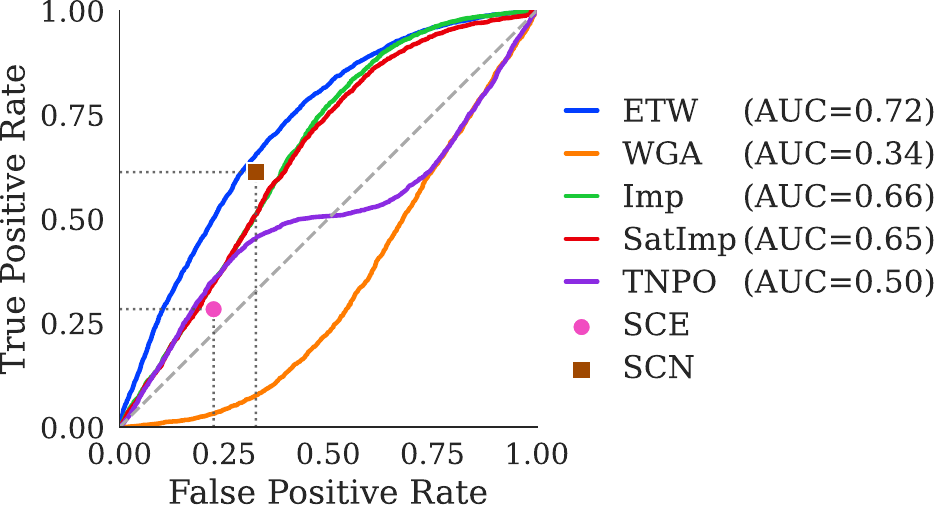}
        \caption{ROC–AUC curves for distinguishing informative tokens from structural tokens across token regularizers on the TOFU forget 10\% split. As SCE and SCN produce binary decisions, only a single operating point (TPR, FPR) is shown. The AUC value for each method is reported in the legend.}
        \label{fig:roc_auc}
\end{figure}
\begin{figure*}[t]
    \centering
    \begin{subfigure}{0.19\linewidth}
        \centering
        \includegraphics[width=\linewidth]{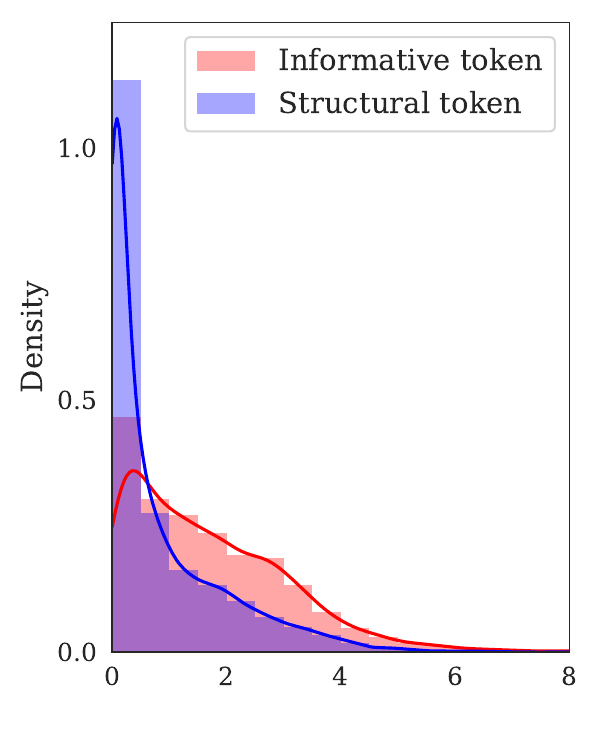}
        \vspace{-22pt}
        \caption{ETW}
        \label{fig:entropy_t15}
    \end{subfigure}
    \begin{subfigure}{0.19\textwidth}
        \centering
        \includegraphics[width=\linewidth]{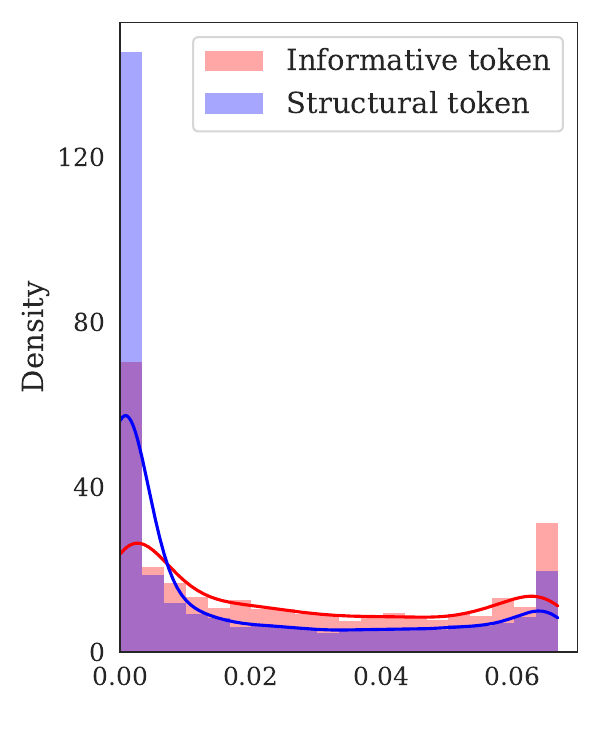}
        \vspace{-22pt}
        \caption{SatImp ($\beta=5.0$)}
        \label{fig:satimp}
    \end{subfigure}
    \begin{subfigure}{0.19\textwidth}
        \centering
        \includegraphics[width=\linewidth]{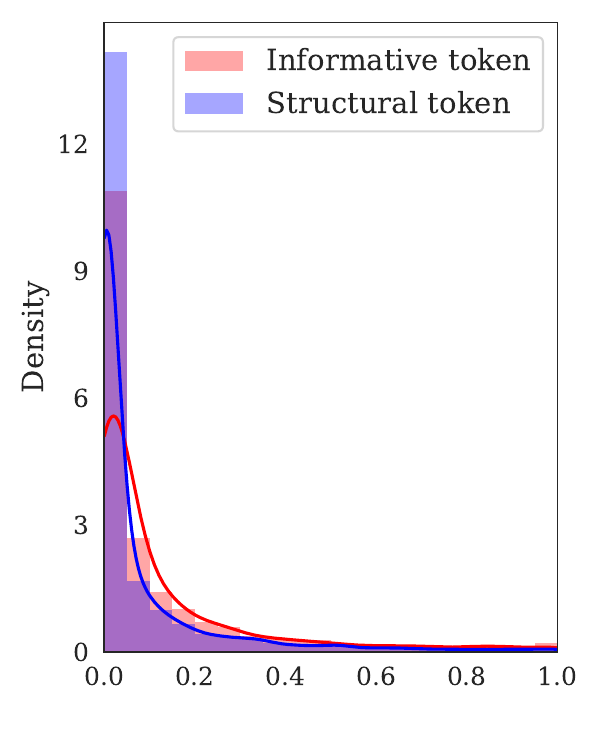}
        \vspace{-22pt}
        \caption{Imp}
        \label{fig:wga}
    \end{subfigure}
    \begin{subfigure}{0.19\textwidth}
        \centering
        \includegraphics[width=\linewidth]{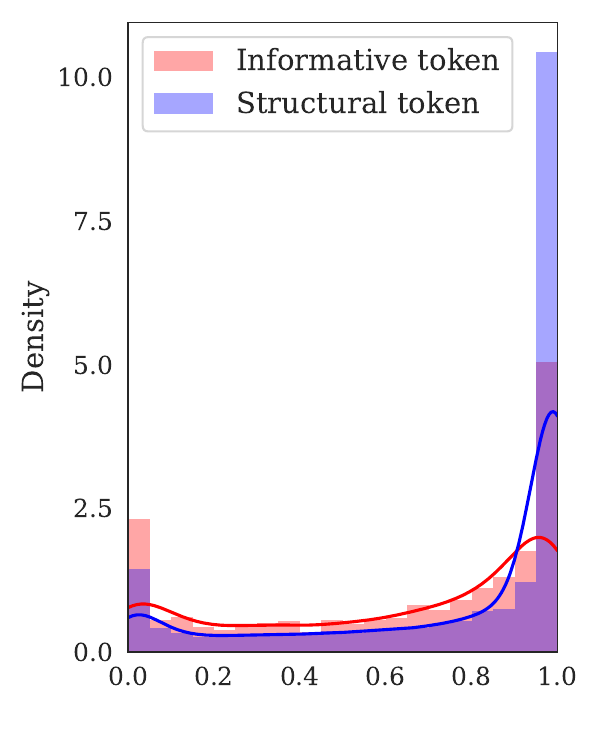}
        \vspace{-22pt}
        \caption{WGA ($\beta=7.0$)}
        \label{fig:imp}
    \end{subfigure}
    \begin{subfigure}{0.19\textwidth}
        \centering
        \includegraphics[width=\linewidth]{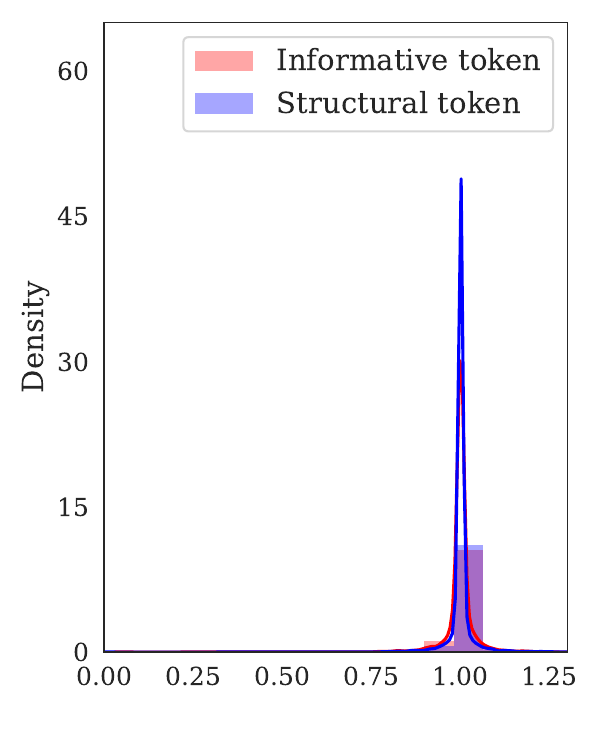}
        \vspace{-22pt}
        \caption{TNPO ($\beta=4.0$)}
        \label{fig:tnpo}
    \end{subfigure}
    \vspace{-2pt}
    \caption{
        Token-wise histograms for informative and structural tokens. We compare ETW with other weighting schemes. The x-axis denotes the computed token weights, while the y-axis represents the probability density. Weights assigned to informative and structural tokens are colored red and blue, respectively.
    }
    \label{fig:histogram}
\end{figure*}

\section{Understanding the Behavior of Token Regularizers}
\label{sec:analysis}

In this section, we provide an in-depth analysis of ETW and existing token regularizers in \Cref{tab:token_regularizers}, focusing on their underlying mechanisms and behavioral differences.

\subsection{Comparative Analysis of ETW and Existing Token Regularizers}
\label{sec:conf_analysis}

Existing token regularizers \cite{wga,satimp,undial} in \Cref{tab:token_regularizers} are categorized as confidence-based and linguistic-based. Confidence-based regularizers, including WGA, Imp, SatImp, and TNPO, commonly penalize tokens associated with specific confidence values. WGA and TNPO penalize tokens with extreme confidence, such as near-deterministic predictions or tokens whose confidence closely matches that of a reference model. Imp attempts to introduce semantic awareness by penalizing the complement of confidence by assigning the strongest penalty to tokens with near-zero confidence. SatImp combines WGA and Imp, aiming to balance confidence suppression and semantic filtering. In fact, its weighting function penalizes tokens most strongly at $p_i = \frac{\alpha}{\alpha + 1}$ (e.g., $\frac{5}{6} \approx 0.83$ for $\alpha = 5.0$). 

These confidence-based regularizers fail to distinguish cases described in \Cref{fig:example}, whereas entropy separates tokens with identical ground-truth confidence. By leveraging the full predictive distribution, ETW provides a significantly richer representational signal than regularizers that rely solely on ground-truth confidence. This allows more accurate measurement of token informativeness.

Linguistic-based approaches such as SpaCy-Noun (SPN) and SpaCy-Entity (SPE) \cite{spacy,undial} adopt spaCy as an external parser to extract nouns or named entities, respectively. While ETW accounts for preceding context and modulates informativeness based on the token prediction distribution, these approaches operate purely at the lexical level and ignore context. As a result, all nouns or entities are treated without considering their contextual relevance or semantic informativeness with respect to the question, and the binary masking scheme further limits fine-grained token-level weighting.

\subsection{Quantitative Evaluation on Informative Token Distinguishability}

\begin{figure*}[t]
    \centering
    \begin{subfigure}[t]{\linewidth}
        \centering
        \includegraphics[
            width=\linewidth,
            trim=0 410 0 0,
            clip
        ]{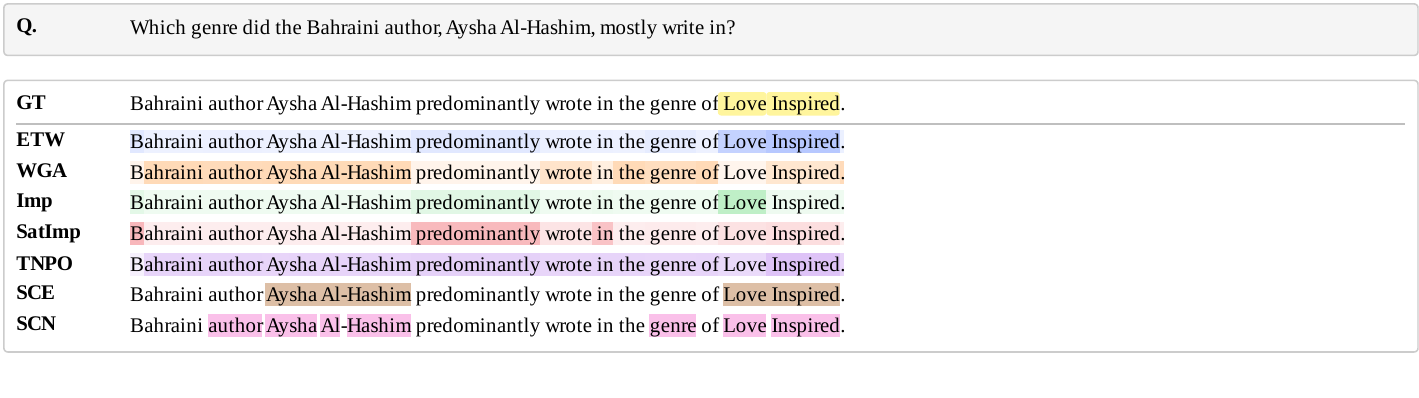}
    \end{subfigure}
    \vspace{-3pt}
    \begin{subfigure}[t]{\linewidth}
        \centering
        \includegraphics[
            width=\linewidth,
            trim=0 420 0 0,
            clip
        ]{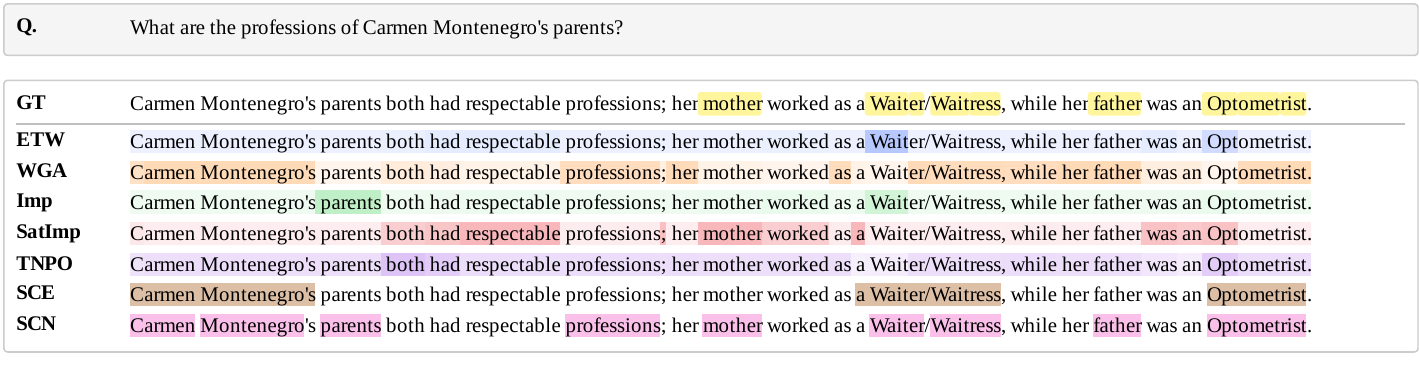}
    \end{subfigure}
    \caption{Token-wise visualization on TOFU forget samples, highlighting informative annotations and forget loss weights for the answer. The forget loss weights are computed using the model before unlearning. Color intensity reflects the degree of regularization of each token. Tokens with higher weights are more strongly highlighted, while lower-weight tokens appear more transparent. Fully transparent tokens are not involved in the unlearning process.}
    \label{fig:highlight}
\end{figure*}

In \Cref{fig:roc_auc}, the performance of each token regularizer in distinguishing informative tokens from structural tokens is evaluated using ROC curves. The important-word annotations introduced by \citet{satimp} for the TOFU dataset are used as ground-truth (GT) labels, where only core answer spans are considered informative and all remaining tokens are treated as structural.

ETW outperforms all other baselines by at least 0.06 in terms of ROC-AUC, demonstrating its superior ability to differentiate informative tokens from structural ones. Among linguistics-based baselines, SCN exhibits relatively strong discriminative performance, but ETW achieves a more favorable TPR–FPR trade-off, attaining lower FPR at the same TPR. For confidence-based methods, Imp performs best with an AUC of 0.66, while SatImp is bounded by the discriminative capability of Imp. In contrast, TNPO and WGA perform at or below random guessing.

The distributions of token regularization weights for informative and structural tokens across different soft regularization methods are shown in \Cref{fig:histogram}. ETW provides a clear separation between informative and structural tokens. The ETW values of structural tokens are concentrated in the low-entropy region, whereas informative tokens exhibit a higher density in the high-entropy regime.

It is notable that the weight distributions in \Cref{fig:satimp,fig:wga,fig:imp} reveal distinct behaviors across confidence-based regularizers. For Imp and SatImp, structural tokens exhibit a higher density near zero weight, while informative tokens occupy a broader overlapping region. This indicates that some informative tokens are also predicted with high confidence and thus receive low weights under Imp and SatImp. Therefore, confidence alone is insufficient for reliably identifying informative content.

WGA applies stronger penalties to high-confidence tokens, in contrast to Imp. It assigns relatively lower weights to informative tokens while concentrating a large portion of structural tokens near confidence 1.0. Likewise, TNPO assigns weights close to 1.0, limiting discrimination between informative and structural tokens.

\subsection{Qualitative Visualization on Token Regularization}
\label{understand qualitative}

\Cref{fig:highlight} presents representative examples in which tokens are colored with varying intensity according to their unlearning weights. ETW exhibits clear discrimination for identifying important tokens, consistent with the quantitative results in \Cref{fig:histogram,fig:roc_auc}. In the upper example, ETW is the only method that successfully identifies the single core answer span, “Love Inspired”. This behavior persists in the lower example, which contains a longer and more complex completion. ETW highlights the key morphemes “Wait” and “Opt”, which form the semantic roots of the parents’ occupations. Methods such as WGA, Imp, and SatImp penalize tokens based on fixed confidence regimes, with WGA and Imp specifically operating in opposing confidence ranges, while TNPO is further constrained by its dependence on a reference model. Consequently, these methods often emphasize words that lack semantic informativeness. SCE and SCN emphasize nouns or named entities regardless of the context. This often causes them to highlight tokens that merely repeat information from the question (e.g., “author” or “Aysha Al-Hashim”), leading to redundant unlearning penalties.

These observations suggest that identifying ``informative'' tokens based solely on part-of-speech categories, entity types, or scalar confidence values is insufficient for capturing truly answer-critical tokens in selective unlearning.

\begin{table*}[t]
\centering
\small
\resizebox{\textwidth}{!}{%
\setlength{\tabcolsep}{1.5pt}
\begin{tabular}{l|rrrr|rrrr|rrrr|rrrr}
\toprule
{Model}
& \multicolumn{4}{c|}{Forget 10\%} 

& \multicolumn{4}{c|}{Forget 05\%}
& \multicolumn{4}{c|}{Forget 01\%} 
& \multicolumn{4}{c}{Avg.} 
\\
 & \scriptsize -log(FQ) $\downarrow$
 & \scriptsize$\Delta$ MU $\downarrow$
 & \scriptsize Agg. $\downarrow$
 & \scriptsize Priv. $|\cdot| \downarrow$
 & \scriptsize -log(FQ) $\downarrow$
 & \scriptsize $\Delta$ MU $\downarrow$
 & \scriptsize Agg. $\downarrow$
 & \scriptsize Priv. $|\cdot| \downarrow$
 & \scriptsize -log(FQ) $\downarrow$
 & $\Delta$ MU $\downarrow$
 & \scriptsize Agg. $\downarrow$
 & \scriptsize Priv. $|\cdot| \downarrow$
 & \scriptsize -log(FQ) $\downarrow$
 & $\Delta$ MU $\downarrow$
 & \scriptsize Agg. $\downarrow$
 & \scriptsize |Priv.| $\downarrow$ \\
 
\toprule
\multicolumn{17}{c}{\textit{Llama 3.2-1B}} \\
\midrule

Ref
 & 20.780 & 0.000 & 0.000 & -99.43
 & 12.184 & 0.000 & 0.000 & -99.98
 & 2.896  & 0.000 & 0.000 & -100.00
 & 11.953 & 0.000 & 0.000 & 99.80 \\

Retrain
 & 0.000	 &1.100 &0.000	 &-0.53	
  &0.000	 &0.447 &	0.000 &	-0.15
 & 0.000 & 0.006 & 0.000 & -2.24
 & 0.000 & 0.518  & 0.000 & 0.97 \\

\midrule
GA
 & 2.639&	4.271&	11.273&	-48.59
 & 3.366&	9.704&	32.658&	\underline{3.07}	
 & 0.238&	11.546&	2.743	&18.65
 & 2.081 & 8.507 & 15.558 & 23.44 \\

SCE
 & 7.971&	3.867&	30.822&	-12.00	
 &0.405 &	9.886 &	4.004 &	\textbf{2.51}
 &\textbf{0.037}&	7.272&	0.267&	-49.47 
 & 2.804 & 7.008 & 11.698 & 21.33 \\

SCN
 & 2.754&	\underline{3.548} &	9.772&	-39.98
 &	1.056&	8.719	&9.211&	-37.37	
 &1.266&	4.928&	6.241	&-63.75 
 & 1.692 & 5.732 &8.408 & 47.03 \\

NPO
 & \underline{1.029} &	6.642&	\underline{6.835}	&-39.15
 &	\textbf{0.101}	&9.591&	\textbf{0.964}&	-19.87	
 &0.238	&7.317&	1.739	& \underline{-10.27} 
 & \underline{0.456} & 7.850 & \underline{3.179} & 23.10 \\

TNPO
 & 4.301&	4.508&	19.390	&-49.09	
 &1.167	&9.269&	10.820&	-21.48	
  &0.116	& 2.162 &	0.25	& 28.22
 & 1.861 & \textbf{5.313} &10.153 & 32.93 \\

WGA
 & 2.309&	5.365&	12.388	&\textbf{-3.14}
 &	0.657&	8.315&	5.459	&-4.00
 &	\textbf{0.037}	&\textbf{3.492}	&\textbf{0.128}&	23.49
 & 1.001 & 5.724 & 5.992  & \underline{10.21} \\

SatImp
 & 2.871&	5.258&	15.093	&-19.65
 &	0.847&	\textbf{8.036}&	6.810&	6.70	
 & \textbf{0.037 }&	\underline{3.548}	& \underline{0.130}	&44.51 
 & 1.252 & \underline{5.614} & 7.344 & 23.62 \\

\midrule
ETW
 & \textbf{0.492}	&\textbf{3.471}&	\textbf{1.707}	& \underline{-9.56}
 & \underline{0.263} &	\underline{8.312} &	\underline{2.189}	&-3.51
 &	\textbf{0.037}	&7.185	&0.264&	\textbf{6.14} 
 & \textbf{0.264} & 6.323 & \textbf{1.387} & \textbf{6.40} \\

\midrule
\midrule

\multicolumn{17}{c}{\textit{Llama 3.2-3B}} \\
\midrule

Ref
 & 27.159&	0.000&	0.000&	-99.75	
 &13.955	&0.000	&0.000&	-100.00
 &1.845&	0.000&	0.000&	-100.00 
 & 14.320 & 0.000 & 0.000 & 99.92 \\

Retrain
 & 0.000&	1.984&	0.000&	-0.39
 & 0.000&	0.454&	0.000&	-0.32	
 & 0.000	&0.478	&0.000	& 1.69 
 & 0.000 & 0.972 & 0.000 & 0.80 \\
\midrule

GA
 & 4.738&	4.431&	20.990&	-40.36
 & 0.950&	8.111&	7.703&	\underline{-8.36}	
 &0.238	&6.052	&1.438	&\textbf{-8.05} 
 & 1.975 & 6.198 & 10.044 & \textbf{18.92} \\

SCE
 & 7.224&	5.343&	38.600&	-32.03
 &	1.403	&6.384&	8.955&	\textbf{-3.25}	
 &\textbf{0.116}	&4.343	&0.503	&-64.27 
 & 2.914 & 5.357 & 16.019 & 33.18 \\

SCN
 & 7.224	&4.959&	35.827	&-57.97
 &	1.403&	\underline{3.781}&	5.304	&-14.34
 &\textbf{0.116}	&6.559	&0.760&	-45.76 
 & 2.914 & 5.100 & 13.964 & 39.36 \\

NPO
 & \underline{1.106} &	4.343	& \underline{4.804}	&-35.38	
 & \textbf{0.016} &	6.355&	\textbf{0.099}&	-20.30
 &1.266	&\textbf{0.022}	&\textbf{0.028}	&-45.20 
 & \underline{0.796} & 3.573 & \underline{1.644} & 33.63 \\

TNPO
 & 5.510&	5.556&	30.613&	\textbf{7.34}
 &	3.015&	7.766	&23.412	&-63.89	
 &\textbf{0.116}	&1.086&	\underline{0.126}	&72.46 
 & 2.880 & 4.803 & 18.050 & 47.90 \\

WGA
 & 6.511	&\textbf{2.581}&	16.805&	\underline{18.81}	
 &1.927&	7.224&	13.917&	-9.96
 &0.393&	\underline{0.744}	&0.292&	-52.68 
 & 2.944 & \underline{3.516} & 10.338 & 27.15 \\
SatImp
 & 8.959&	\underline{3.242} &	29.044&	19.13
 &	6.094&	\textbf{0.981}&	5.978&	41.36	
 &0.238	&3.487	&0.829	& \underline{18.93} 
 & 5.097 & \textbf{2.570} & 11.950 & 26.47 \\
\midrule

ETW
 & \textbf{0.330} & 3.652 & \textbf{1.204} & -18.96
 & \underline{0.263} & 5.925 & \underline{1.561} & -18.40
 & \textbf{0.116} & 4.588 & 0.531 & -8.89
 & \textbf{0.236} & 4.722 & \textbf{1.099} & \underline{15.42} \\

\midrule\midrule

\multicolumn{17}{c}{\textit{Llama 2-7B}} \\
\midrule

Ref
 & 24.703 & 0.000 & 0.000 & -99.87
 & 12.876 & 0.000 & 0.000 & -100.00
 & 2.896 & 0.000 & 0.000 & -100.00
 & 13.492 & 0.000 & 0.000 & 99.96 \\

Retrain
 & 0.000 & 2.329 & 0.000 & -0.36
 & 0.000 & 0.172 & 0.000 & -0.37
 & 0.000 & 0.219 & 0.000 & 1.89
 & 0.000 & 0.907 & 0.000 & 0.87 \\

\midrule

GA
 & 6.169 & 1.325 & 8.175 & -25.74
 & 1.167 & 2.668 & 3.114 & -14.45
 & 2.170 & 0.772 & 1.675 & -65.11
 & 3.169 & 1.588 & 4.321 & 35.10 \\

SCE
 & 5.672 & \textbf{-2.133} &\textbf{ -12.099} & -12.16
 & 12.876 & \textbf{0.539} & 6.944 & 12.15
 & \underline{0.238} & 6.160 & 1.464 & 34.26
 & 6.262 & 1.522 & \textbf{-1.230} & 19.52 \\

SCN
 & 8.361 &\underline{ 0.101} & \underline{0.848} & -73.54
 & 2.681 & 4.251 & 11.397 & \textbf{-1.76}
 & 2.896 & \underline{-0.279} & \textbf{-0.808} & -100.00
 & 4.646 & \textbf{1.358} & 3.812 & 58.43 \\

NPO
 & \textbf{0.330} & 6.330 & 2.087 & -15.42
 & \textbf{0.101} & 2.607 & \textbf{0.262} & -7.04
 & \textbf{0.116} & 1.728 & 0.200 & \textbf{2.39}
 & \textbf{0.182} & 3.555 & \underline{0.850} & 8.28 \\

TNPO
 & 2.527 & 3.202 & 8.092 & -29.35
 & 0.484 &\underline{ 1.845} & \underline{0.893} & -22.47
 & 1.013 & -0.158 & -0.160 & 82.62
 & 1.341 & 1.630 & 2.942 & 44.81 \\

WGA
 & 7.224 & 1.655 & 11.955 & 46.73
 & 1.927 & 2.770 & 5.336 & 13.35
 & 1.266 & \textbf{-0.300} & \underline{-0.380} & 96.35
 & 3.472 & \underline{1.375} & 5.637 & 52.14 \\

SatImp
 & 4.160 & 2.549 & 10.605 & \underline{2.96}
 & 1.056 & 2.956 & 3.123 & \underline{2.74}
 & 0.393 & 1.104 & 0.434 & \underline{-13.35}
 & 1.870 & 2.203 & 4.721 & \textbf{6.35} \\

\midrule

ETW
 & \underline{1.999} & 1.816 & 3.630 & \textbf{0.76}
 & \underline{0.263} & 4.338 & 1.143 & -4.09
 & 1.266 & 0.668 & 0.846 & 14.86
 & \underline{1.176} & 2.274 & 1.873 & \underline{6.57} \\

 \bottomrule
\end{tabular}
}
\caption{Experiments on the TOFU benchmark with 10\%, 5\%, and 1\% forget splits on LLaMA-3.2 1B, 3B, and LLaMA 2-7B. The best results are \textbf{bolded}, and the second-best are \underline{underlined}. For Avg. $-\log(\mathrm{FQ})$, $\Delta$MU, and Agg., values are averaged, while Priv. is averaged in absolute terms (|Priv.|, $\downarrow$).}
\label{tab:tofu_main}
\end{table*}

\section{Experiment}
\subsection{Experimental Setting}
We conduct experiments on the TOFU benchmark \cite{tofu} using the
10\%, 5\%, and 1\% forget splits with LLaMA-3.2 \cite{llama} models of sizes 1B and 3B and WMDP benchmark \cite{wmdp} with StableLM Zephyr 3B\cite{zephyr}.
Our implementation is based on the \texttt{open-unlearning}
repository \cite{openunlearning}.
TOFU experiments are trained for 10 epochs, and WMDP experiments are trained for 125 epochs. All experiments are conducted on a single NVIDIA A6000 GPU,
using a batch size of 2 with gradient accumulation over 8 steps,
resulting in an effective batch size of 16.
We use a learning rate of $1\times10^{-5}$ with a linear scheduler.

GA corresponds to unlearning with the GA loss without token-wise weighting. NPO \cite{npo} is trained using the NPO loss. The remaining variants apply the token regularizers in \Cref{tab:token_regularizers} to \Cref{eq:weighted_ga}.
For the TOFU experiment, the reference (Ref) model is fine-tuned on both the forget and retain sets, while the retrained (Retrain) model is trained using only the retain set. 

\subsection{Metrics}

\paragraph{TOFU Benchmark}

We evaluate unlearning performance using forget quality (FQ, $\uparrow$) and model utility (MU, $\uparrow$) from the TOFU benchmark.
For reporting, we use $-\log(\mathrm{FQ})$ ($\downarrow$), while model utility is measured as the relative degradation from the reference model:
\begin{equation}
\Delta \mathrm{MU}(\cdot) =
\frac{\mathrm{MU(Ref)} - \mathrm{MU}(\cdot)}{\mathrm{MU(Ref)}} \times 100 (\%) .
\end{equation}

We then introduce an aggregated metric to identify the best configuration that achieves both high forget quality and high model utility.
Since $-\log(\mathrm{FQ})$ and $\Delta\mathrm{MU}$ are comparably scaled, we report an aggregated metric (Agg., $\downarrow$) defined as their product:
\begin{equation}
\mathrm{Agg.} = \left(-\log(\mathrm{FQ})\right) \times \Delta \mathrm{MU}.
\end{equation}
This multiplicative form penalizes configurations that perform poorly on either criterion, thereby favoring balanced trade-offs between forgetting quality and utility preservation.
This metric is used to select the best-performing model in \Cref{tab:tofu_main}. To avoid trivial near-zero scores arising from undesirable cases, we exclude out-of-range configurations from the best-model selection procedure.
Specifically, we discard models with $-\log(\mathrm{FQ}) > 12$ or $\Delta\mathrm{MU} < 50\%$, which correspond to cases where forget quality shows insufficient improvement over the reference model or model utility is severely degraded.

We additionally report Privleak (Priv., $|\cdot| \downarrow$) as an auxiliary privacy-related metric \cite{muse,mink}, where values near zero indicate balanced unlearning, positive values over-unlearning, and negative values under-unlearning.

\paragraph{WMDP Benchmark} 
For the WMDP benchmarks, multiple-choice datasets are used for evaluation; therefore, accuracy is reported. WMDP-Cyber and WMDP-Bio are used to measure forget-set accuracy, while the MMLU QA set \cite{mmlu} is used to evaluate retain-set accuracy.
To ensure fair comparison, we select each baseline such that its MMLU accuracy is close to 35\%, thereby aligning retaining performance across methods.

\subsection{Experiments on TOFU Benchmark}
\label{sec:satimp_exp}

\Cref{tab:tofu_main} presents the results on the TOFU benchmark, comparing all methods across multiple forget splits and model sizes. In the 1B and 3B settings, ETW achieves the best forget quality while preserving model utility, resulting in the lowest average aggregated score (Agg. in Avg.) across forget splits. In the 7B setting, ETW maintains a competitive aggregated score without collapsing in utility, demonstrating stable performance across model sizes. This advantage becomes more pronounced as the forget rate increases, where most baselines suffer significant degradation while ETW remains robust. This gap is particularly evident under the 10\% forget setting. ETW also demonstrates competitive privacy leakage performance across model sizes, achieving the lowest privacy leakage among all baselines for the 1B model.

Since forget quality is defined as the p-value of a Kolmogorov--Smirnov test, a value greater than 0.05 indicates that the output distributions of the unlearned and retrained models are not statistically distinguishable \cite{scipy}, corresponding to $-\log(\mathrm{FQ})<1.3$. Therefore, in the 10\% forget split for the 1B and 3B models, only ETW and NPO demonstrate meaningful unlearning in terms of forget quality. Notably, under the 5\% forget setting for 1B and 3B models, ETW exhibits the smallest loss in model utility among methods that satisfy the forget-quality criterion, leading to a competitive aggregated score. At the 1\% forget split, where only 2 out of 200 authors are removed, most methods achieve acceptable forget quality. In this easier regime, achieving forget quality requires less unlearning pressure, allowing methods that better preserve utility to appear competitive. This is in contrast to the 10\% forget setting, where the task is substantially harder and ETW's ability to balance forget quality and utility preservation is most evident.

The 7B results further highlight the scalability of ETW. Although NPO achieves the strongest forget quality in the 7B setting, ETW maintains competitive overall performance as measured by the aggregated score while preserving valid unlearning across most forget splits. In the 10\% forget split, SCE achieves a lower aggregated score, but this is driven by $\Delta\mathrm{MU}<0$, indicating a slight increase in model utility after unlearning, while its $-\log(\mathrm{FQ})$ substantially exceeds the validity threshold of 1.3.

\begin{table}[t]
\centering
\setlength{\tabcolsep}{2pt}
\renewcommand{\arraystretch}{0.9}
\resizebox{\linewidth}{!}{%

\begin{tabular}{l r r r r r r r}
\toprule
Model
& \multicolumn{2}{c}{Forget 10\%}
& \multicolumn{2}{c}{Forget 05\%}
& \multicolumn{2}{c}{Forget 01\%} 
& \multicolumn{1}{c}{Avg.}\\
&  Prob. &  $\lvert \Delta$RT $\rvert$
&  Prob. &  $\lvert \Delta$RT $\rvert$
&  Prob. &  $\lvert \Delta$RT $\rvert$
&  $\lvert \Delta$RT$\rvert$ \\
\midrule
Ref
 & 0.957 & --
 & 0.953 & --
 & 0.953 & -- \\

Retrain
&	0.415&	0.0&
	0.384	&0.0&	
0.399	&0.0 
& 0.0 \\

\midrule
GA
& 	0.578&	39.2&
0.309&	19.5	&
0.339	& \underline{14.9} 
& 24.5 \\

SCE
&0.647	&55.8
&0.532 &	38.5	
&0.602	& 50.9
& 48.4 \\

SCN
&	0.539&	29.7
&	0.486&	26.6	
&0.577&	44.5 
& 33.6 \\

NPO
&	0.568&	36.6
&	0.526& 36.9	
&	0.561&	40.6 
& 38.0 \\

TNPO
&	0.595&	43.1
&	0.517&	34.6	
&0.971	& 143.3 
& 73.7 \\

WGA
&	0.422&	\underline{1.6} &
0.499&	29.9&	
0.489	& 22.5 
& \underline{18.0} \\

SatImp
&	0.499	& 20.2
	&0.444&	\underline{15.5}
&0.472&	18.3 
& \underline{18.0} \\

\midrule
ETW
& 0.415&	\textbf{0.2}&
0.365&	\textbf{5.0}&
0.418&	\textbf{4.7} 
& \textbf{3.3} \\

\bottomrule
\end{tabular}
}
\caption{Probability (Prob.) on informative tokens on the TOFU dataset for LLaMA-3.2-1B  with 10\%, 5\%, and 1\% forget splits. $\lvert \Delta$RT $\rvert$(\%) denotes the absolute relative change with respect to the Retrain model, $\lvert$ $\frac{(\text{Retrain}-x)}{\text{Retrain}}$ $\rvert \times 100 $(\%). The smallest $\lvert \Delta \mathrm{RT} \rvert$ values are \textbf{bolded}, and the second-best are \underline{underlined}.}
\label{tab:forget_prob_delta}
\end{table}

\subsection{Probability Analysis of Informative Tokens}
\label{sec:prob_anlaysis_satimp}
We further investigate whether unlearned models successfully avoid assigning high probability to important tokens by analyzing the token-level probabilities (Prob.) of informative tokens on the TOFU dataset using LLaMA-3.2 1B.
We focus on how closely these probabilities align with those of the retrained model.
Across all forget splits, ETW exhibits the smallest probability gap relative to the retrained model, with at most a 5\% difference.
Notably, under the 10\% forget setting, the difference is as small as 0.2\%.

In contrast, GA tends to overly suppress informative tokens under the 5\% and 1\% settings, which is consistent with its overall utility degradation observed in \Cref{tab:tofu_main}.
SCN and TNPO, on the other hand, fail to sufficiently forget informative tokens and produce probabilities close to those of the reference model.
These results indicate that strong forget quality scores alone do not necessarily imply effective suppression of informative tokens.

\subsection{Temperature analysis}

\begin{figure}[t]
\centering
        \includegraphics[width=\linewidth]{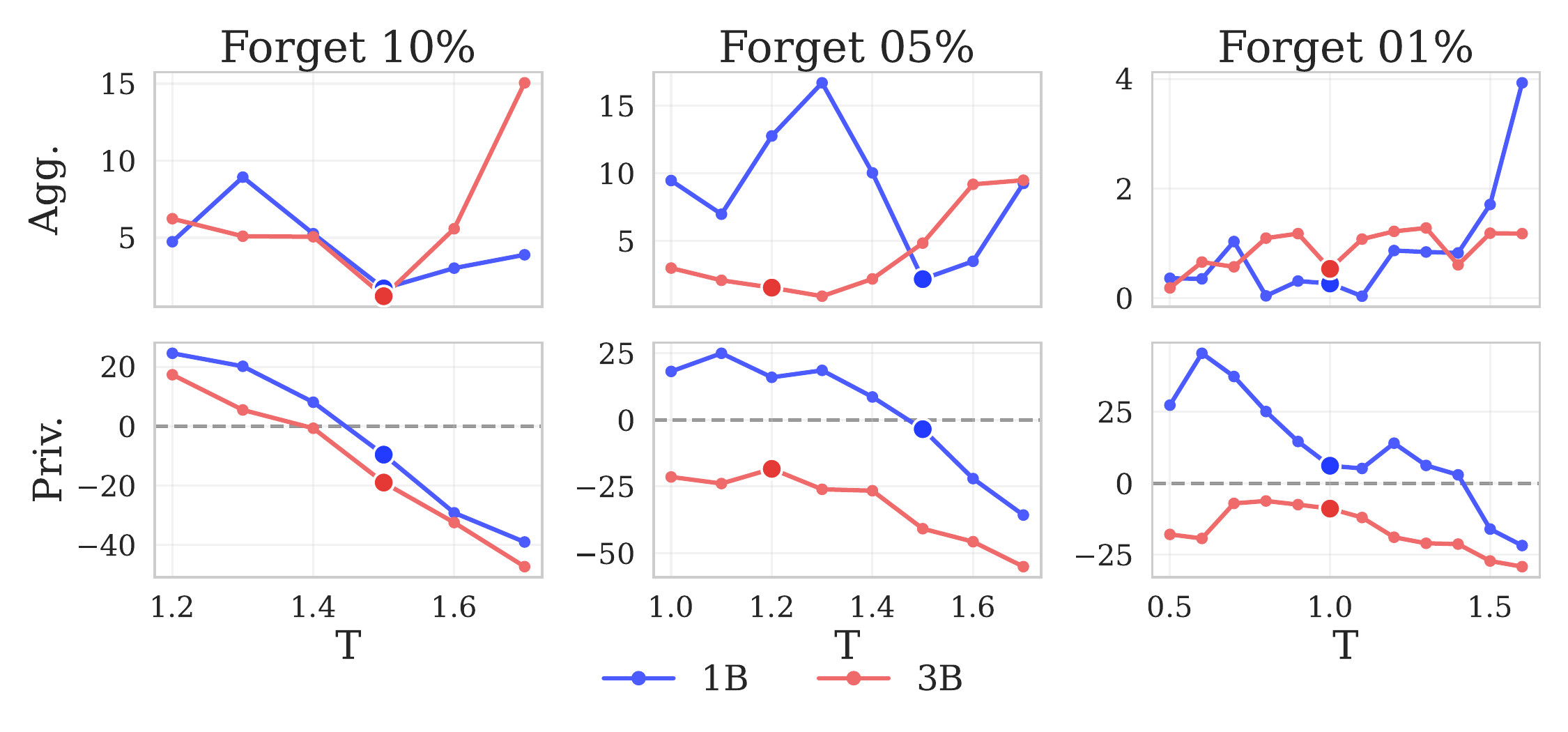}
        \caption{Aggregated score (Agg.) and privacy leakage (Priv.) on the TOFU dataset across temperature values and forget splits of 10\%, 5\%, and 1\% for LLaMA 3.2 1B and 3B models. The configuration used in \Cref{tab:tofu_main} is highlighted with larger circle markers.}
        \label{fig:temp_main}
\end{figure}

In \Cref{fig:temp_main}, we report aggregated score (Agg.) and privacy leakage (Priv.) across temperature settings for LLaMA 3.2 1B and 3B models on the TOFU dataset, along with model utility degradation ($\Delta$MU) and forget quality ($-\log(\mathrm{FQ}$)) in \Cref{fig:temperature_supple} in the Appendix. Temperature influences both the extent of unlearning within each split and the optimal choice across splits. This sensitivity to temperature arises because temperature controls the sharpness of the probability distribution and, in turn, the strength of entropy-based penalization, as shown in \Cref{fig:entropy_temperature} in the Appendix.

Within a given split, privacy leakage generally decreases as temperature increases, transitioning from over-unlearning to under-unlearning. Across splits, the optimal temperature depends on the number of forget samples. Larger splits with sufficient data favor a moderate penalty induced by higher temperatures. In contrast, smaller splits with limited samples benefit from stronger penalization at lower temperatures.

\subsection{Experiments on WMDP Benchmark}
 
For the WMDP benchmarks (\Cref{tab:wmdp_exp}), we report models with similar MMLU accuracies. Accordingly, their retaining performance exhibits little variation, and there is also little variation in WMDP-Cyber accuracy. ETW records one of the lowest accuracies on WMDP-Bio, indicating a successful trade-off in unlearning effectiveness.

While the multiple-choice nature of WMDP does not evaluate fine-grained performance, ETW demonstrates a more favorable trade-off than NPO by yielding lower accuracy on both forget sets. Although SCE shows better forgetting performance by achieving lower Cyber accuracy at the same Bio accuracy level, ETW consistently exhibits robust performance across both TOFU and WMDP benchmarks, highlighting its general effectiveness.

\begin{table}[t]
\centering
\small
\begin{tabular}{l | c c |c}
\toprule
\multirow{2}{*}{Model} & \multicolumn{2}{c|}{Forget (Acc, $\%$) ($\downarrow$ )} & \multicolumn{1}{c}{Retain (Acc, $\%$) ($\uparrow$)} \\
& Cyber (\(\downarrow\)) & Bio (\(\downarrow\)) & MMLU (\(\uparrow\)) \\
\midrule
Ref          & 35.78 & 53.02 & 45.04 \\
\midrule
GA           & \underline{29.79} & 39.83 &  35.13 \\
SCE          & \textbf{28.64} & \textbf{36.45} & 35.06 \\
SCN          & 31.25 & 38.57 & 34.30 \\
NPO          & 30.05 & 37.31 & 35.29 \\
TNPO         & 29.94 & 40.06 & \textbf{35.87} \\
WGA          & 29.89 & 38.65 & \textbf{36.16} \\
SatImp       & 30.95 & 40.61 & 35.27 \\
\midrule
ETW          & \underline{29.79} & \textbf{36.45} & 35.22 \\
\bottomrule
\end{tabular}
\caption{Results on the WMDP-Cyber and WMDP-Bio benchmarks on StableLM Zephyr 3B. All metrics are reported as accuracy (\%). For the forget sets, lower accuracy is better. For the retain set, higher accuracy is better. The best accuracies are \textbf{bolded}, and the second-best are \underline{underlined}.}
\label{tab:wmdp_exp}
\end{table}

\section{Conclusion}

In this paper, we introduce Entropy-guided Token Weighting (ETW), a token-level regularizer that leverages the entropy of the predictive distribution as a proxy for token informativeness to enhance selective unlearning in LLMs. ETW applies stronger penalties to high-entropy informative tokens while preserving low-entropy structural tokens, effectively addressing the limitations of prior methods that rely on confidence or external linguistic parsers. ETW outperforms existing baselines in token discrimination, and models unlearned with ETW exhibit prediction patterns for informative tokens that are similar to those of retrained models. By selectively penalizing informative content, experiments on the TOFU and WMDP benchmarks demonstrate that ETW achieves a superior trade-off between knowledge forgetting and preserving model utility.

\section*{Acknowledgements}
This work was partly supported by the Institute of Information \&
Communications Technology Planning \& Evaluation (IITP) grant funded by
the Korea government (MSIT) (No.RS-2025-02283048, Developing the Next-Generation General AI with Reliability, Ethics, and Adaptability, 80\%) and Basic Science Research Program through the National Research Foundation of Korea (NRF) funded by the Ministry of Education (No.RS-2025-25410835, Machine Unlearning in Continual Learning with Linearity-Based Reduction of Data Dependency for Trustworthy AI, 20\%).

\section*{Limitations}
Our study has several limitations. First, our unlearning objective focuses on gradient ascent loss. While it is a widely used and effective unlearning approach, different unlearning objectives may exhibit distinct optimization dynamics. As a result, applying ETW to other unlearning objectives could lead to different stability and efficiency characteristics.

Second, token-wise regularization cannot fully mitigate side effects arising from the auto-regressive nature of LLMs. Although informative tokens are penalized more strongly during training, this only modifies token-level loss terms. At inference time, small changes in early token predictions can alter the entire generation trajectory, leading to cascading effects that are difficult to anticipate, and making a comprehensive analysis of all generation paths infeasible.

Third, the behavior of token distributions may vary across different model architectures. As a result, the effectiveness and characteristics of ETW may differ when applied to models with substantially different predictive dynamics.

\bibliography{custom}

\appendix


\clearpage
\noindent
\begin{minipage}{\textwidth}
\centering
\small
\setlength{\tabcolsep}{4pt}
\resizebox{\textwidth}{!}{%
\begin{tabular}{l |c c| c c| c c | c c| c c| c c | c c| c c| c c}
\toprule
\multirow{2}{*}{Model} 
& \multicolumn{6}{c|}{\textbf{Llama 3.2-1B}}
& \multicolumn{6}{c|}{\textbf{Llama 3.2-3B}}
& \multicolumn{6}{c}{\textbf{Llama 2-7B}} \\
\cmidrule(lr){2-7}\cmidrule(lr){8-13}\cmidrule(lr){14-19}
& \multicolumn{2}{c|}{Forget 10\%}
& \multicolumn{2}{c|}{Forget 05\%}
& \multicolumn{2}{c|}{Forget 01\%} 
& \multicolumn{2}{c|}{Forget 10\%}
& \multicolumn{2}{c|}{Forget 05\%}
& \multicolumn{2}{c|}{Forget 01\%}
& \multicolumn{2}{c|}{Forget 10\%}
& \multicolumn{2}{c|}{Forget 05\%}
& \multicolumn{2}{c}{Forget 01\%}\\
& $\lambda$ & Param.
& $\lambda$ & Param.
& $\lambda$ & Param. 
& $\lambda$ & Param.
& $\lambda$ & Param.
& $\lambda$ & Param.
& $\lambda$ & Param.
& $\lambda$ & Param.
& $\lambda$ & Param. \\
\midrule
GA     & 0.10 & --                & 0.15 & --                & 0.50 & -- 
       & 0.09 & --                & 0.15 & --                & 0.30 & -- 
       & 0.09 & --                & 0.10 & --                & 0.10 & -- \\
SCE    & 0.04 & --                & 0.08 & --                & 0.30 & -- 
       & 0.05 & --                & 0.10 & --                & 0.30 & -- 
       & 0.05 & --                & 0.05 & --                & 0.30 & -- \\
SCN    & 0.06 & --                & 0.07 & --                & 0.20 & -- 
       & 0.05 & --                & 0.10 & --                & 0.40 & -- 
       & 0.05 & --                & 0.07 & --                & 0.01 & -- \\
NPO    & 200  & $\beta{=}0.5$     & 600  & $\beta{=}0.5$     & 20   & $\beta{=}0.5$ 
       & 300  & $\beta{=}0.5$     & 1000 & $\beta{=}0.5$     & 0.20 & $\beta{=}0.5$ 
       & 300  & $\beta{=}0.5$                 & 600  & $\beta{=}0.5$  & 100  & $\beta{=}0.5$   \\
TNPO   & 1    & $\beta{=}4$       & 1    & $\beta{=}4$       & 3    & $\beta{=}4$ 
       & 0.5  & $\alpha{=}4$      & 0.5  & $\alpha{=}4$      & 5    & $\alpha{=}4$ 
       & 0.4  & $\alpha{=}4$ & 0.7  & $\alpha{=}4$ & 2    & $\alpha{=}4$ \\
WGA    & 5    & $\alpha{=}7$      & 3    & $\alpha{=}5$      & 4    & $\alpha{=}5$ 
       & 10   & $\alpha{=}7$      & 1    & $\alpha{=}5$      & 2    & $\alpha{=}5$
       & 20   & $\alpha{=}7$ & 1    & $\alpha{=}5$ & 3    & $\alpha{=}5$ \\
SatImp & 15   & $\alpha{=}5$      & 20   & $\alpha{=}5$      & 40   & $\alpha{=}5$ 
       & 150  & $\alpha{=}5$      & 150  & $\alpha{=}5$      & 70   & $\alpha{=}5$
       & 7    & $\alpha{=}5$ & 10   & $\alpha{=}5$ & 30   & $\alpha{=}5$ \\
\midrule
ETW    & 0.06 & $T{=}1.5$         & 0.06 & $T{=}1.5$         & 0.15 & $T{=}1.0$ 
       & 0.045 & $T{=}1.5$        & 0.045 & $T{=}1.2$        & 0.07 & $T{=}1.0$
       & 0.025 & $T{=}1.2$ & 0.028 & $T{=}1$ & 0.03 & $T{=}0.7$ \\
\bottomrule
\end{tabular}
}
\captionof{table}{Best hyperparameter configurations for each method under different TOFU forget splits.}
\label{tab:best_hyperparams}
\end{minipage}
\section{External Package Specification}
For the spaCy baselines, we use spaCy v3.8.7 with the \texttt{en\_core\_web\_sm} model (v3.8.0). We compute ROUGE scores using the \texttt{rouge-score} library (v0.1.2),
a pure Python implementation of ROUGE-1.5.5.

\section{Benchmarks}
\subsection{TOFU Benchmark}
The TOFU benchmark \cite{tofu} provides a collection of 200 fictitious authors, each associated with 20 question–answer pairs, enabling controlled evaluation of forgetting specific identities. The benchmark supports 10\%, 5\%, and 1\% splits between forget and retain sets. To evaluate retaining performance, TOFU reports model utility, defined as the harmonic mean of several metrics measured on the retain set, including token-wise probability, ROUGE recall, and Truth Ratio, evaluated across retain data, real-author prompts, and world-fact queries. In addition, TOFU proposes a forget quality metric, defined as the p-value of a Kolmogorov–Smirnov (KS) test between the output distributions of a model retrained solely on the retain set and an unlearned model. This metric is computed by comparing the Truth Ratio distributions of the two models, providing a statistical measure of how closely the unlearned model approximates retraining from scratch. Membership inference attack metrics, such as MinK \cite{mink}, have been increasingly adopted as measures of privacy preservation in LLM unlearning, as demonstrated in \cite{muse}.

\subsection{WMDP Benchmark}
The WMDP benchmark \cite{wmdp} consists of 3,668 multiple-choice questions designed to assess hazardous knowledge in biosecurity (1,273), cybersecurity (1,987), and chemical security (408). In this paper, we conduct unlearning experiments on the cybersecurity (WMDP-Cyber) and biosecurity (WMDP-Bio) subsets. Unlearning performance is evaluated by measuring accuracy on these subsets to assess forgetting, while knowledge retaining is measured using MMLU accuracy.

\section{Hyper Parameter Configuration}
\subsection{TOFU Experiment}

In \Cref{tab:best_hyperparams}, we summarize the hyperparameter configurations across different models and forget splits. Following prior work \cite{wga,satimp}, we set $\alpha=5.0$ for all SatImp models, $\beta=4.0$ for TNPO, and $\beta=0.5$ for NPO. For WGA, $\alpha$ is set to $7.0$ under the TOFU 10\% forget split and to $5.0$ for the other settings. For ETW, we additionally tune the softmax temperature and select the best-performing value.

\begin{figure*}[t]
    \centering
    \begin{subfigure}{0.32\linewidth}
        \centering
        \includegraphics[width=\linewidth]{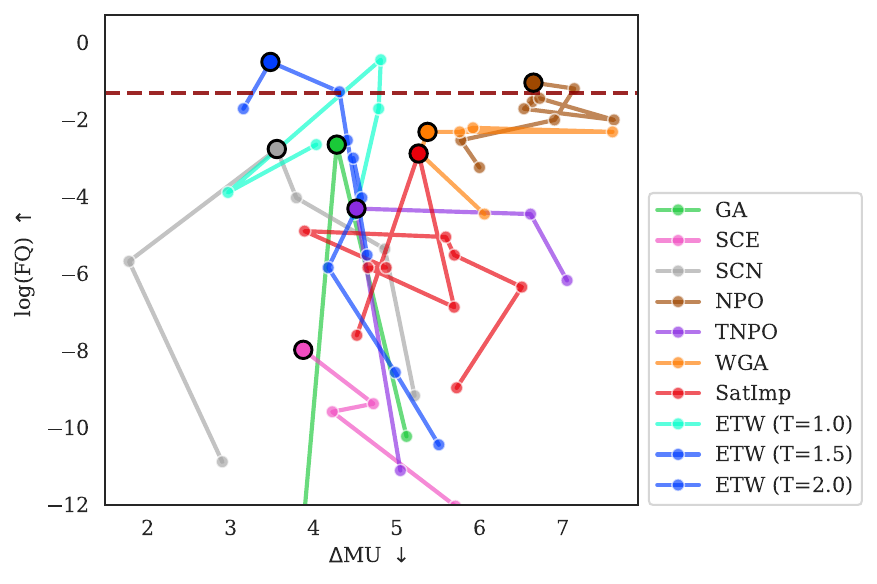}
        \caption{1B with Forget 10\%}
    \end{subfigure}
    \begin{subfigure}{0.32\textwidth}
        \centering
        \includegraphics[width=\linewidth]{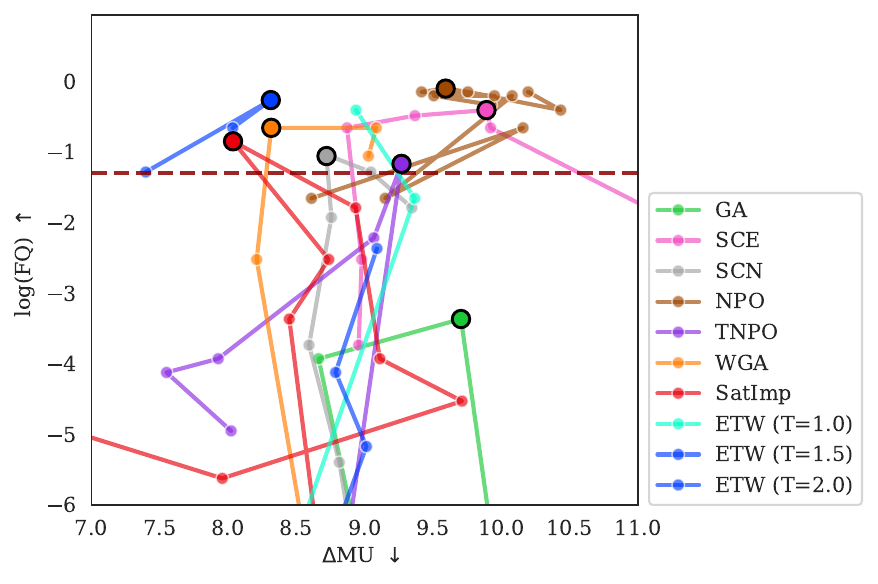}
        \caption{1B with Forget 05\%}
    \end{subfigure}
    \begin{subfigure}{0.32\textwidth}
        \centering
        \includegraphics[width=\linewidth]{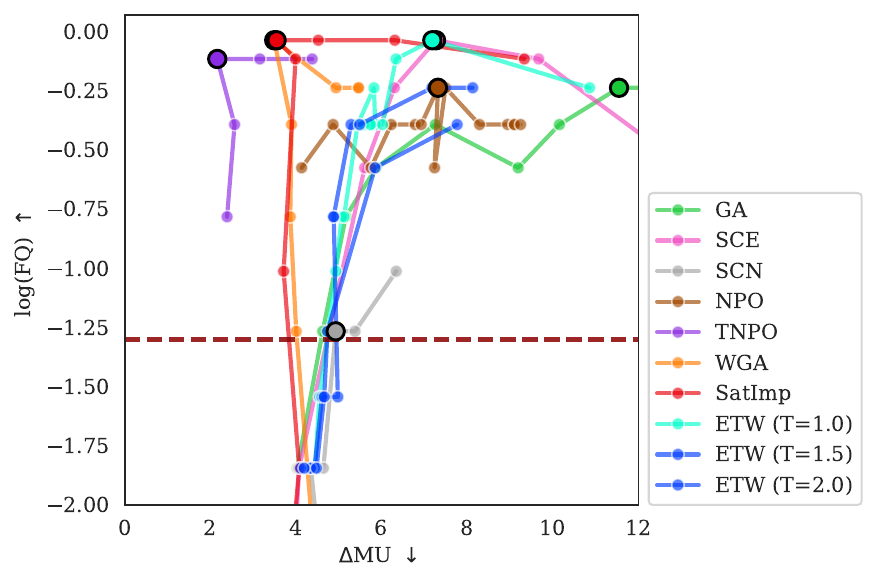}
        \caption{1B with Forget 01\%}
    \end{subfigure}
    
    \begin{subfigure}{0.32\textwidth}
        \centering
        \includegraphics[width=\linewidth]{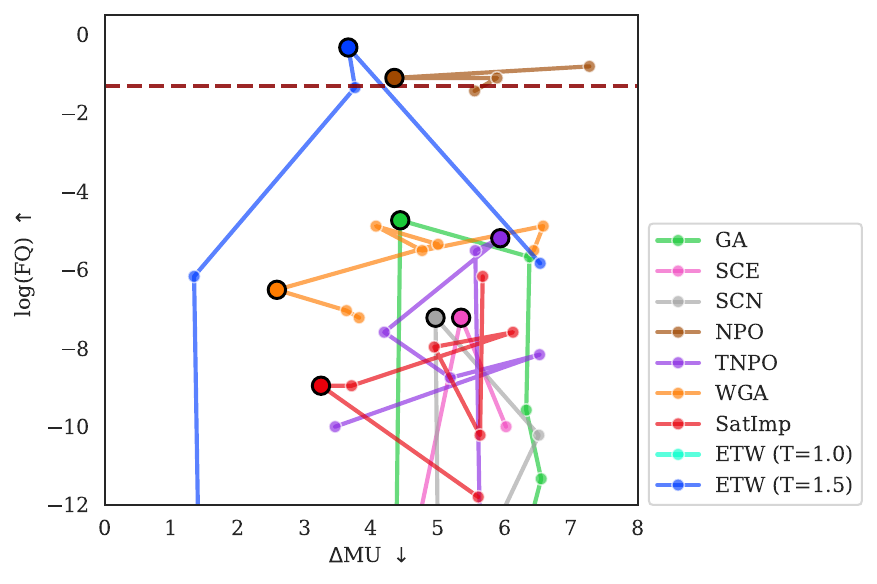}
        \caption{3B with Forget 10\%}
    \end{subfigure}
    \begin{subfigure}{0.32\textwidth}
        \centering
        \includegraphics[width=\linewidth]{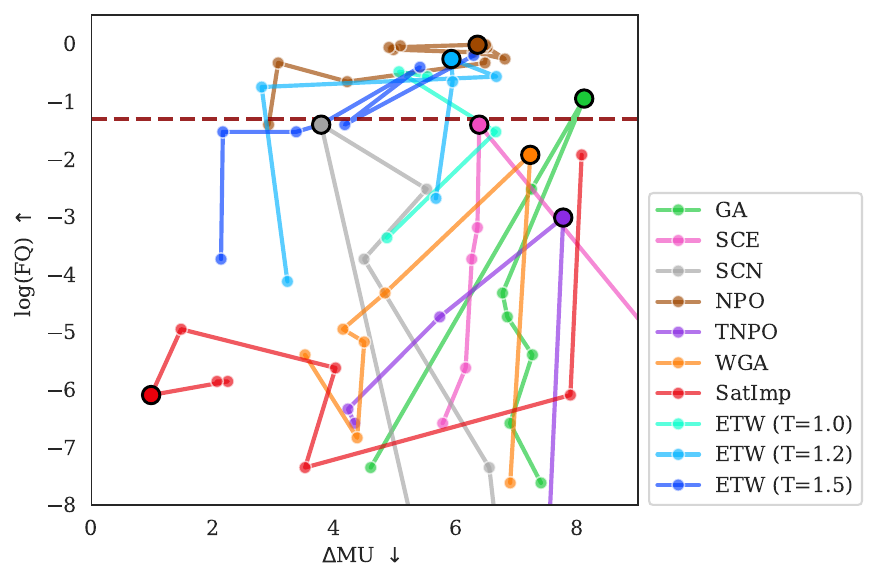}
        \caption{3B with Forget 05\%}
    \end{subfigure}
    \begin{subfigure}{0.32\textwidth}
        \centering
        \includegraphics[width=\linewidth]{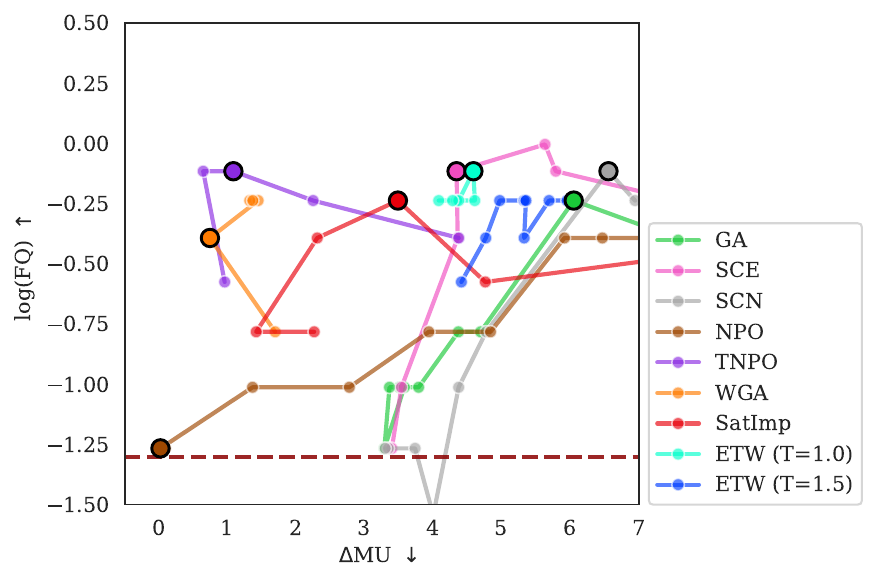}
        \caption{3B with Forget 01\%}
    \end{subfigure}
    \vspace{0.5em}
    \caption{Trade-off between model utility and forget quality on TOFU.
Larger markers indicate the best configuration for each method.
The upper-left region, with higher $\log(\mathrm{FQ})$ and lower $\Delta\mathrm{MU}$, represents better unlearning–retention trade-offs.
    }
    \label{fig:hyperparam}
\end{figure*}

To obtain these configurations, we search for the optimal $\lambda$ for each model size and forget split by balancing high forget quality (larger $\log(\mathrm{FQ})$) and low utility degradation (smaller $\Delta\mathrm{MU}$), as illustrated in \Cref{fig:hyperparam}.

\subsection{WMDP Experiment}
In \Cref{tab:wmdp_benchmark}, we report the hyperparameter configuration used to achieve approximately 35\% MMLU accuracy.

\begin{table}[t]
\vspace{140pt}
\centering
\begin{tabular}{l c l}
\toprule
Method & $\lambda$ & Param. \\
\midrule
GA     & 0.01 & -- \\
SCE    & 0.02 & -- \\
SCN    & 0.01 & -- \\
NPO    & 0.10 & $\beta = 0.5$ \\
TNPO   & 0.10 & $\beta = 5$ \\
WGA    & 1.00 & $\alpha = 5$ \\
SatImp & 11.0 & $\alpha = 5$ \\
\midrule
ETW    & 0.03 & $T = 1.2$ \\
\bottomrule
\end{tabular}
\caption{Hyperparameters for each unlearning method.}
\label{tab:wmdp_benchmark}
\end{table}

\begin{table}[t]
\centering
\scriptsize
\setlength{\tabcolsep}{2pt}
\resizebox{\linewidth}{!}{%
\begin{tabular}{l r r r r r r r}
\toprule
Model
& \multicolumn{2}{c}{Forget 10\%}
& \multicolumn{2}{c}{Forget 05\%}
& \multicolumn{2}{c}{Forget 01\%} 
& \multicolumn{1}{c}{Avg.}\\
&  Prob. &  $\lvert \Delta$RT $\rvert$
&  Prob. &  $\lvert \Delta$RT $\rvert$
&  Prob. &  $\lvert \Delta$RT $\rvert$
&  $\lvert \Delta$RT$\rvert$ \\
\midrule

Ref
 & 0.985 & -
 & 0.983 & -
 & 0.988 & -
 & -\\

Retrain
 & 0.422 & 0.0
 & 0.392 & 0.0
 & 0.419 & 0.0 
 & 0.0\\

\midrule
GA
 & 0.579 &	37.1
& 0.402	 &\textbf{2.5}
 & 0.453 &	\textbf{8.0} 
 & \underline{15.9} \\

NPO
 & 0.543 & 28.6
& 0.539 & 37.4
 & 0.649 & 54.8 
 & 40.3 \\

TNPO
 & 0.436 & \textbf{3.4}
& 0.619 & 57.8
 & 0.375 & 10.6 
 & 23.9 \\

WGA
 & 0.366 & 13.4
& 0.439 & 11.9
 & 0.603 & 43.8 
 & 23.0 \\

SatImp
 & 0.362 & 14.3
 & 0.286 & 27.1
& 0.483 & 15.1 
 & 18.8\\

SCE
& 0.693 & 64.2
& 0.544 & 38.7
& 0.557 & 32.9 
& 45.3 \\

SCN
 & 0.602 & 42.6
 & 0.377 & \underline{3.8}
 & 0.475 & 13.4 
 & 19.9 \\

\midrule
ETW
 & 0.471 & \underline{11.4}
 & 0.414 & 5.7
 & 0.460 & \underline{9.8} 
 & \textbf{9.0}\\

\bottomrule
\end{tabular}
}
\caption{Probability (Prob.) on informative tokens for LLaMA-3.2-3B under different TOFU forgetting ratios. $\lvert \Delta$RT $\rvert$(\%) denotes the absolute relative change with respect to the Retrain model, $\lvert$ $\frac{(\text{Retrain}-x)}{\text{Retrain}}$ $\rvert \times 100 $(\%). The smallest $\lvert \Delta$RT $\rvert$ are \textbf{bolded}, and the second-best are \underline{underlined}.}
\label{tab:forget_prob_delta_3b}
\end{table}

\section{Additional Result of Probability Analysis for Mitigating Informative Tokens}

In line with \Cref{tab:forget_prob_delta}, we conduct an analysis on LLaMA-3.2-3B to observe token-wise probabilities of informative tokens. We measure the relative probability gap between each method and the retrained model using $\lvert \Delta \text{RT} \rvert$ in  \Cref{tab:forget_prob_delta_3b}. It is clear that ETW shows the smallest average probability gap of 9.0, followed by GA at 15.9 and TNPO at 23.9. While TNPO and GA achieve the best performance only at specific settings (10\% and 5\% splits for TNPO, and 1\% for GA), ETW consistently maintains low probability gaps across all settings with token-wise probabilities for informative tokens closely matching those of the retrained model.

\begin{figure}[t]
    \centering
    \begin{subfigure}{0.4\linewidth}
        \centering
        \includegraphics[width=\linewidth]{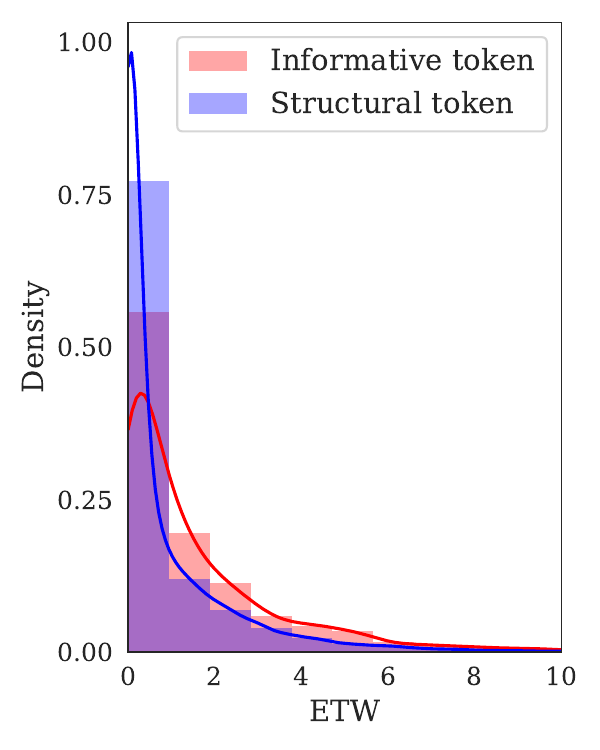}
        \caption{T=1}
        \label{fig:entropy_t10}
    \end{subfigure}
    \begin{subfigure}{0.4\linewidth}
        \centering
        \includegraphics[width=\linewidth]{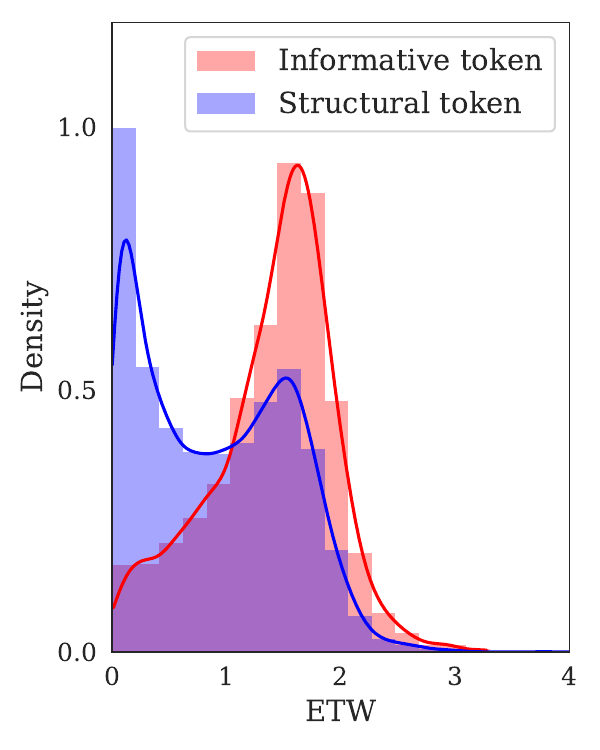}
        \caption{T=2}
        \label{fig:entropy_t20}
    \end{subfigure}
    \caption{Token-wise histogram for informative and structural tokens of ETW with different softmax temperature $T=1.0, T=2.0$.}
    \label{fig:entropy_temperature}
\end{figure}

\section{ETW Distribution under Different Softmax Temperatures}
We use a softmax temperature of $T=1.5$ for the Forget 10\% and Forget 5\% settings,
and $T=1.0$ for the Forget 1\% and WMDP settings.
To examine the effect of temperature on token regularization,
we compare the regularization patterns under $T=1.0$ and $T=2.0$, as shown in \Cref{fig:entropy_temperature}.

\section{Forget quality and model utility across different temperatures}

\begin{figure}[t]
\centering
        \includegraphics[width=\linewidth]{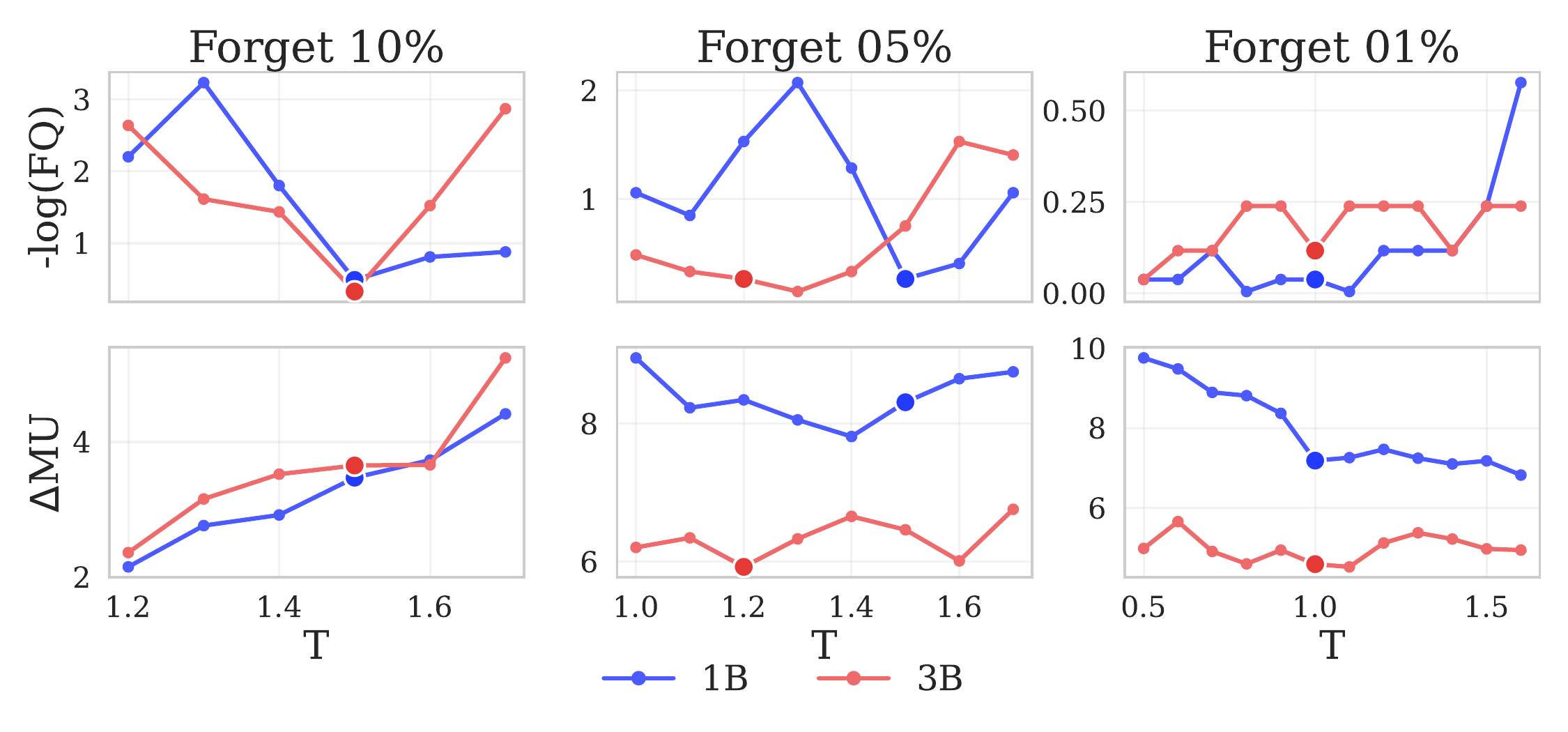}
        \caption{Forget quality ($-\log(\mathrm{FQ})$) and model utility degradation ($\Delta$MU) across different temperatures and forget splits for LLaMA 3.2-1B and 3B models. The configuration used in \Cref{tab:tofu_main} is highlighted with larger circle markers.}
        \label{fig:temperature_supple}
\end{figure}

In \Cref{fig:temperature_supple}, we additionally present the trajectories of forget quality ($-\log(\mathrm{FQ})$) and model utility degradation ($\Delta \mathrm{MU}$), where the aggregated score (Agg.) in \Cref{fig:temp_main} is defined as their product. While $\Delta \mathrm{MU}$ shows relatively moderate sensitivity to temperature in certain settings, such as Forget 5\% for both models and Forget 1\% with the 3B model, $-\log(\mathrm{FQ})$ exhibits a clearer temperature-dependent optimum. As a result, the temperature selected in the main experiments is primarily driven by forget quality.

\section{Epoch-wise Analysis of Token Regularization}

\begin{figure*}[t]
    \centering
    \begin{subfigure}{0.19\linewidth}
        \centering
        \includegraphics[width=\linewidth]{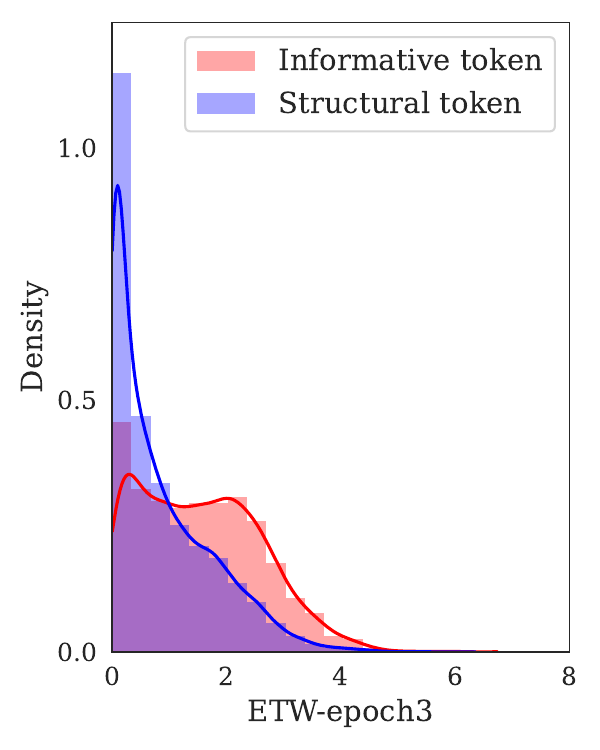}
        \caption{ETW Epoch 3}
        \label{fig:etw 3}
    \end{subfigure}
    \begin{subfigure}{0.19\textwidth}
        \centering
        \includegraphics[width=\linewidth]{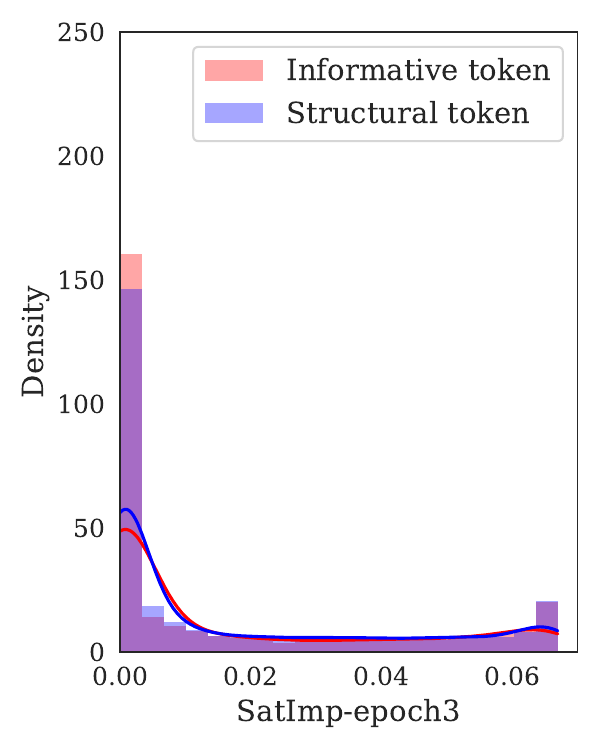}
        \caption{SatImp Epoch 3}
        \label{fig:satimp 3}
    \end{subfigure}
    \begin{subfigure}{0.19\textwidth}
        \centering
        \includegraphics[width=\linewidth]{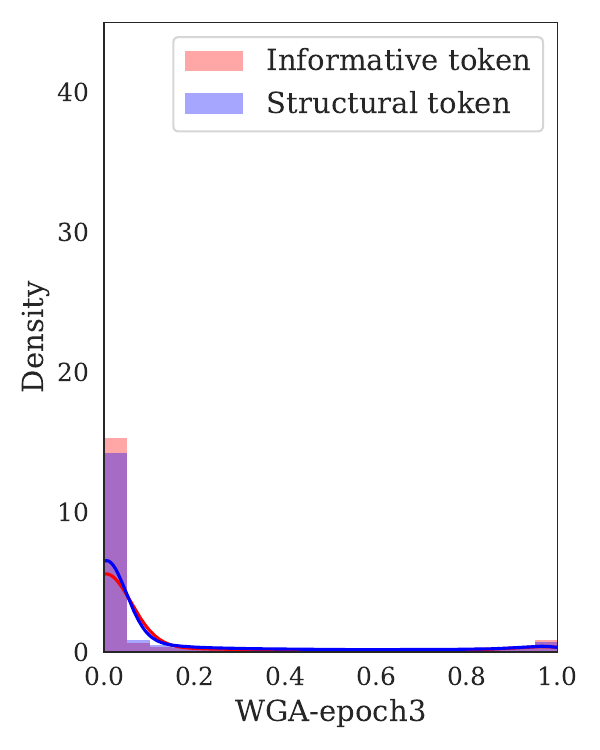}
        \caption{WGA Epoch 3}
        \label{fig:wga 3}
    \end{subfigure}
    \begin{subfigure}{0.19\textwidth}
        \centering
        \includegraphics[width=\linewidth]{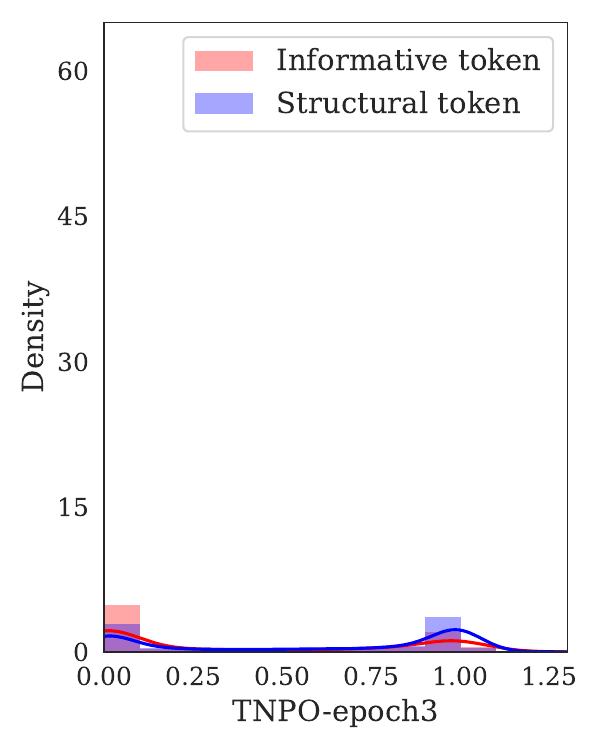}
        \caption{TNPO Epoch 3}
        \label{fig:tnpo 3}
    \end{subfigure}

    \begin{subfigure}{0.19\linewidth}
        \centering
        \includegraphics[width=\linewidth]{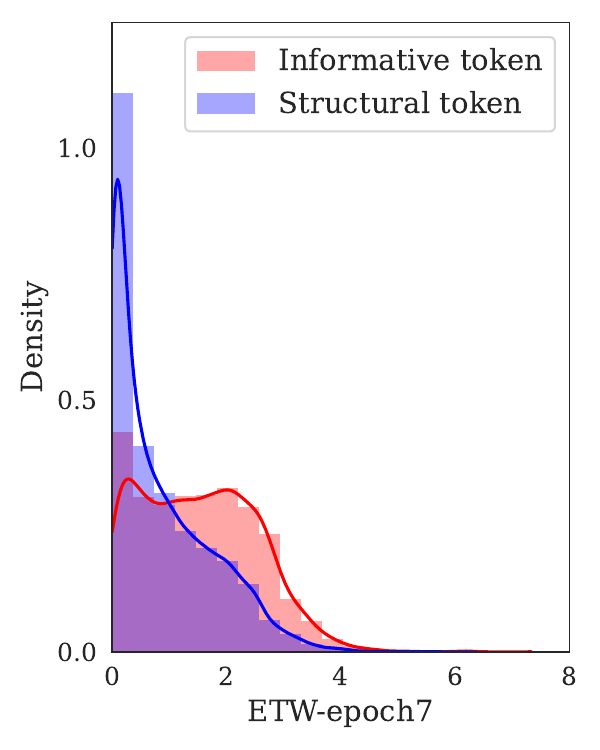}
        \caption{ETW Epoch 7}
        \label{fig:etw 7}
    \end{subfigure}
    \begin{subfigure}{0.19\textwidth}
        \centering
        \includegraphics[width=\linewidth]{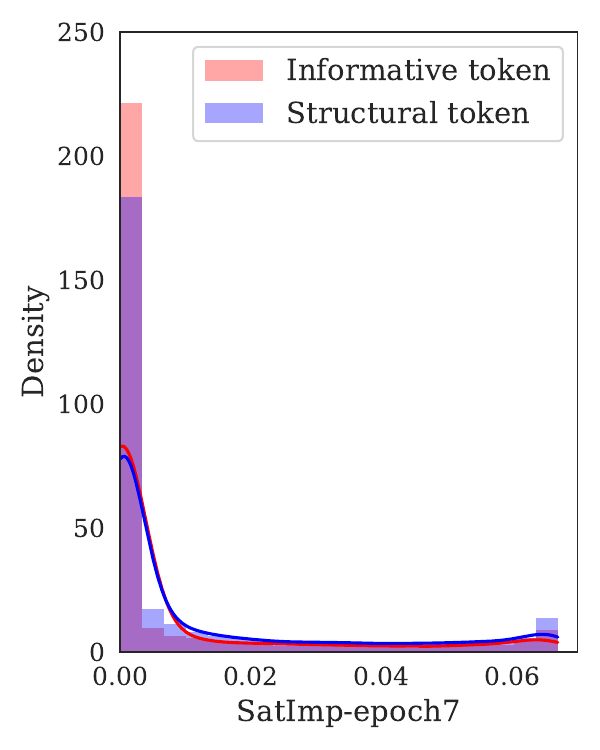}
        \caption{SatImp Epoch 7}
        \label{fig:satimp 7}
    \end{subfigure}
    \begin{subfigure}{0.19\textwidth}
        \centering
        \includegraphics[width=\linewidth]{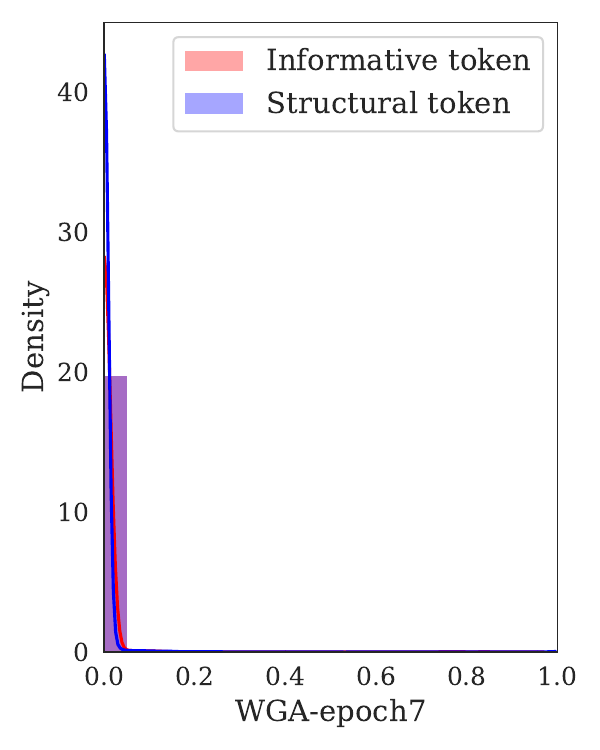}
        \caption{WGA Epoch 7}
        \label{fig:wga 7}
    \end{subfigure}
    \begin{subfigure}{0.19\textwidth}
        \centering
        \includegraphics[width=\linewidth]{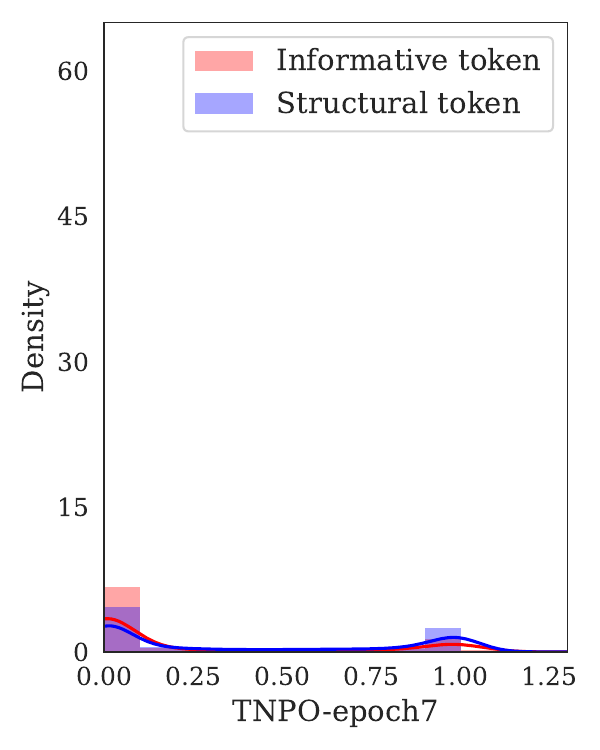}
        \caption{TNPO Epoch 7}
        \label{fig:tnp 7}
    \end{subfigure}

    \begin{subfigure}{0.19\linewidth}
        \centering
        \includegraphics[width=\linewidth]{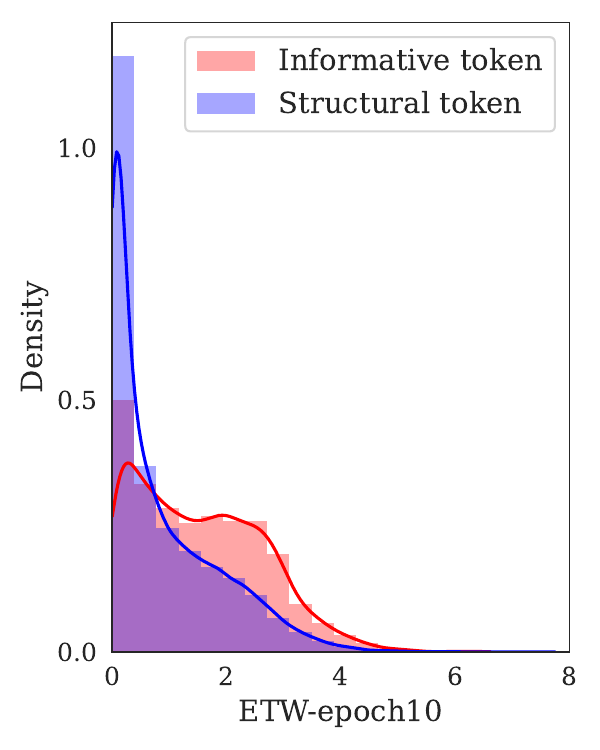}
        \caption{ETW Epoch 10}
        \label{fig:etw 10}
    \end{subfigure}
    \begin{subfigure}{0.19\textwidth}
        \centering
        \includegraphics[width=\linewidth]{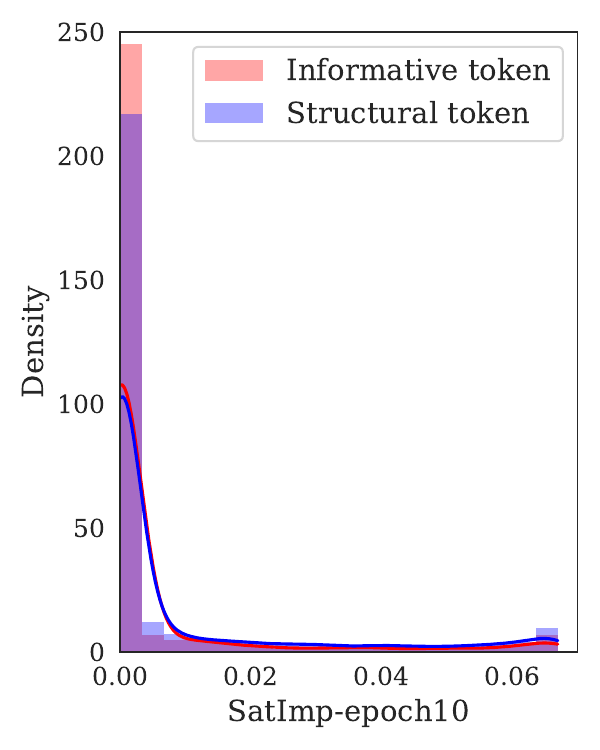}
        \caption{SatImp Epoch 10}
        \label{fig:satimp 10}
    \end{subfigure}
    \begin{subfigure}{0.19\textwidth}
        \centering
        \includegraphics[width=\linewidth]{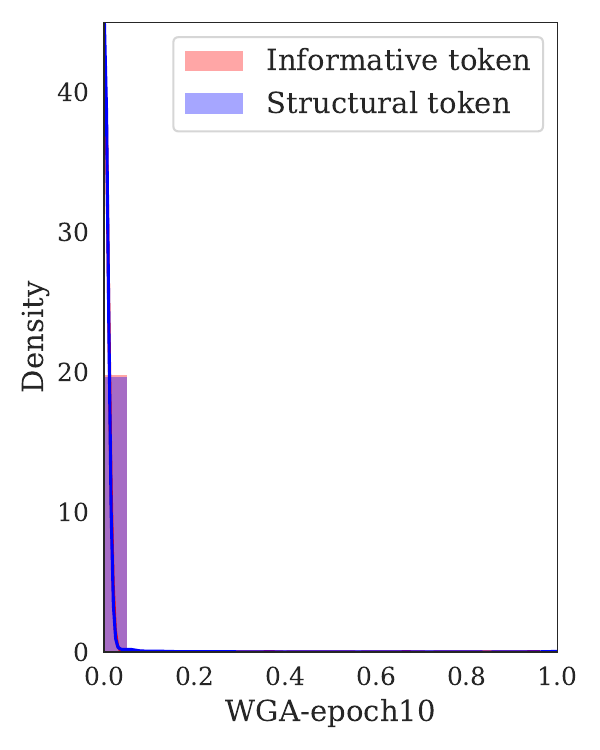}
        \caption{WGA Epoch 10}
        \label{fig:wga 10}
    \end{subfigure}
    \begin{subfigure}{0.19\textwidth}
        \centering
        \includegraphics[width=\linewidth]{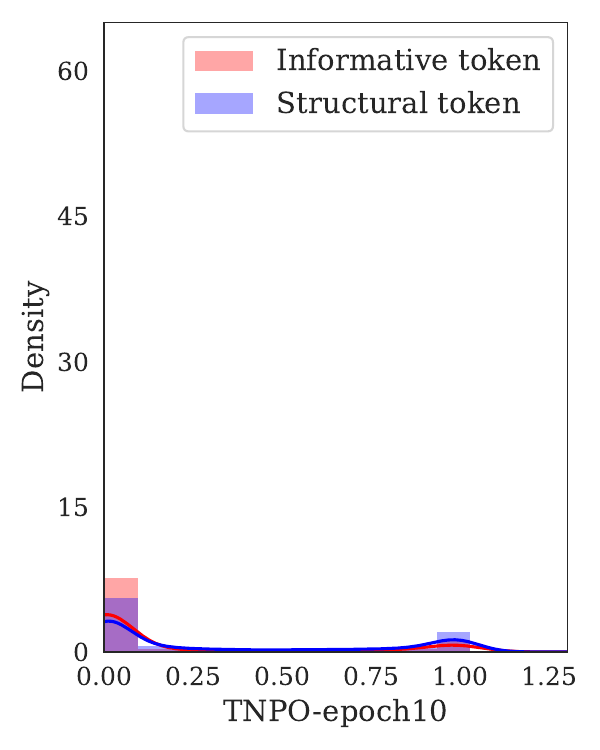}
        \caption{TNPO Epoch 10}
        \label{fig:tnp 10}
    \end{subfigure}

    \begin{subfigure}{0.19\linewidth}
        \centering
        \includegraphics[width=\linewidth]{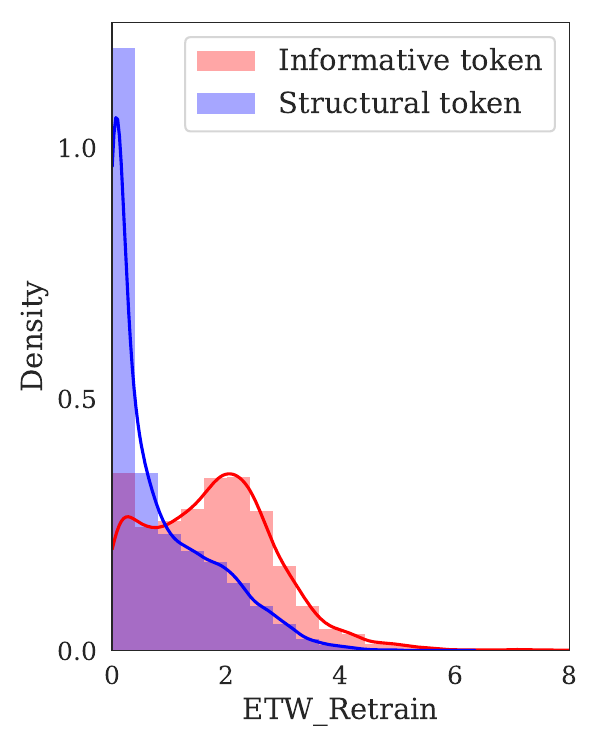}
        \caption{ETW Retrain}
        \label{fig:etw retrain}
    \end{subfigure}
    \begin{subfigure}{0.19\textwidth}
        \centering
        \includegraphics[width=\linewidth]{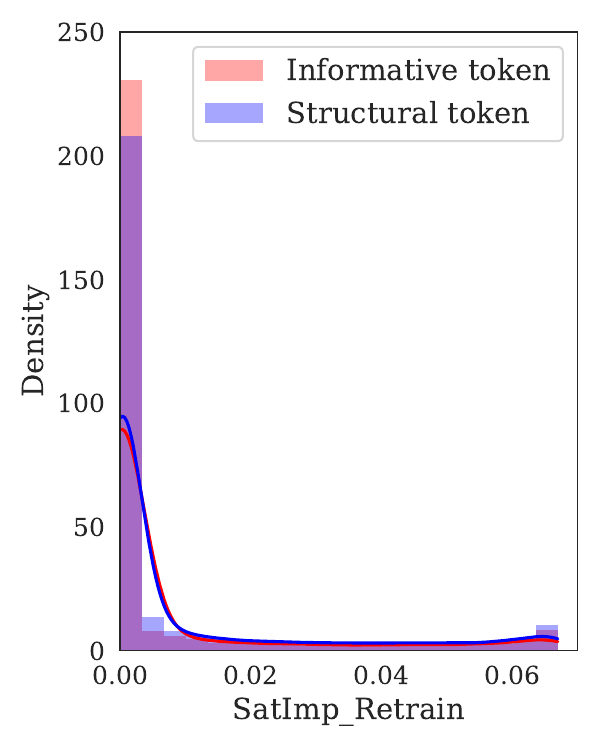}
        \caption{SatImp Retrain}
        \label{fig:satimp retrain}
    \end{subfigure}
    \begin{subfigure}{0.19\textwidth}
        \centering
        \includegraphics[width=\linewidth]{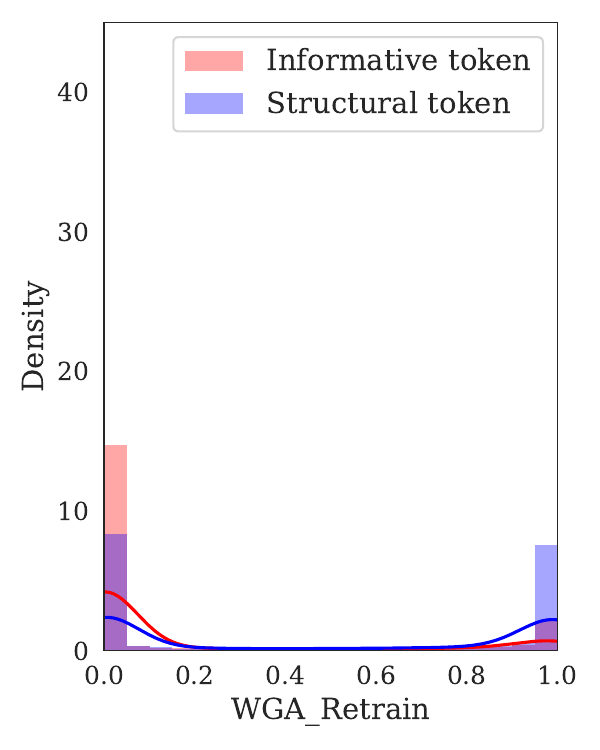}
        \caption{WGA Retrain}
        \label{fig:wga retrain}
    \end{subfigure}
    \begin{subfigure}{0.19\textwidth}
        \centering
        \includegraphics[width=\linewidth]{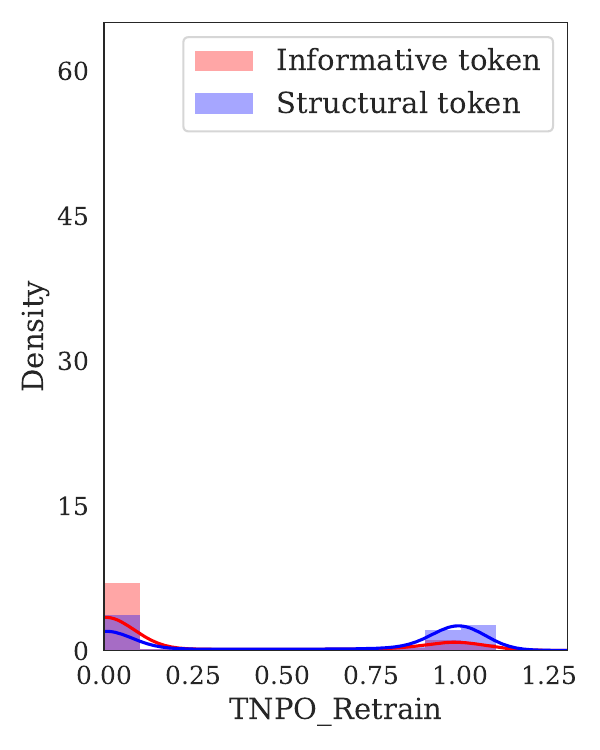}
        \caption{TNPO Retrain}
        \label{fig:tnpo retrain}
    \end{subfigure}
    \vspace{0.5em}
    \caption{Epoch-wise token regularization under TOFU 10\% forget split. Token-level weights of ETW, SatImp, WGA, and TNPO are shown at epochs 3, 7, and 10 for the best models in the 10\% forget split. The last row corresponds to the retrained model trained on the retain set only.
    }
    \label{fig:epoch histogram}
\end{figure*}

We conduct additional analysis to observe how the token-wise weights histogram changes during the unlearning process. Using the best-performing models under the TOFU Forget 10\% setting, we track epoch-wise changes of our proposed method (ETW), alongside SatImp, WGA, and TNPO. Additionally, we report the token-wise weight distributions generated by the retrained model using each method's weighting scheme. While \Cref{fig:histogram} in \Cref{understand qualitative} shows the histogram of the initial model before unlearning, \Cref{fig:epoch histogram} illustrates the intermediate stages during the unlearning process.

As shown in \Cref{fig:etw 3}, \Cref{fig:etw 7}, and \Cref{fig:etw 10}, ETW exhibits similar token-wise weight patterns throughout the unlearning process. This indicates that the model preserves its general language modeling capabilities by applying minimal penalties to structural tokens. Furthermore, since ETW does not induce excessive entropy reduction for informative tokens, the model successfully preserves its informative-token discrimination ability during the unlearning process. Notably, \Cref{fig:etw retrain} and \Cref{fig:etw 10} show that the weight distributions of our final unlearned model show similar patterns to those of the retrained model.

In contrast, other methods exhibit different weight distribution patterns between the final unlearned model and the retrained model. 
WGA shows a notable discrepancy between its Epoch 10 distribution and the retrained model's distribution, as shown in \Cref{fig:wga 10} and \Cref{fig:wga retrain}, where informative tokens are concentrated in lower-weight regions compared to the retrained model. TNPO also places informative tokens disproportionately in lower-weight regions relative to the retrained model, as seen in \Cref{fig:tnp 10} and \Cref{fig:tnpo retrain}, indicating insufficient emphasis on informative content.
In contrast, SatImp maintains relatively similar distributions between Epoch 10 and the retrained model in \Cref{fig:satimp 10} and \Cref{fig:satimp retrain}, as its importance weighting based solely on ground-truth token confidence partially captures token informativeness.

Through epoch-wise distribution analysis, we confirm that our approach preserves the model's ability to distinguish informative tokens while achieving effective unlearning. These weight distribution patterns align with our probability-based analysis in \Cref{sec:prob_anlaysis_satimp}.

\section{Preprocessing for Extracting Ground-Truth Informative Tokens}

\begin{table*}[t]
\centering
\small
\setlength{\tabcolsep}{4pt}
\renewcommand{\arraystretch}{1.15}
\begin{tabular}{r l l p{0.72\textwidth}}
\toprule
{IDX} & {Case} & {Field} & {Content} \\
\midrule

\multirow{4}{*}{0} & \multirow{4}{*}{Normal} & Question &
What is the full name of the author born in Taipei, Taiwan on 05/11/1991 who writes in the genre of leadership? \\
& & Answer &
The author's full name is \underline{\textbf{Hsiao Yun-Hwa}}. \\
& & SatImp &
\texttt{authorTaipeiTaiwan05/11/1991leadershipHsiaoYun-Hwa} \\
& & rule-based-out &
\texttt{['Hsiao Yun-Hwa']} \\
\midrule

\multirow{4}{*}{13} & \multirow{4}{*}{Normal} & Question &
How has her LGBTQ+ identity played a role in the reception of Hsiao Yun-Hwa's leadership books? \\
& & Answer &
Her identity as an LGBTQ+ individual has made Hsiao Yun-Hwa a role model for \underline{\textbf{diverse authors}} and \underline{\textbf{leaders}}. Her perspective has brought a \underline{\textbf{fresh}} and\underline{\textbf{ welcome view }}to leadership literature. \\
& & SatImp &
\texttt{diverseauthorsleadersfreshwelcomeview} \\
& & Rule-based &
\texttt{['diverse authors', 'leaders', 'fresh', 'welcome view']} \\
\midrule

\multirow{4}{*}{5} & \multirow{4}{*}{Typo} & Question &
Can you name an example of Hsiao Yun-Hwa's work that is influenced by her life experiences? \\
& & Answer &
One of Hsiao Yun-Hwa's books, \underline{\textbf{``The Immutable Laws of Engineering Leadership: }} \underline{\textbf{A Blueprint''}}, was noticeably influenced by her \underline{\textbf{father}}'s work as a civil engineer, exhibiting a \underline{\textbf{deep understanding}} of leadership in technical fields. \\
& & SatImp&
\makecell[l]{\texttt{TheImmutbleLawsofEngineerLeadershipABlueprintfatherdeepunderstanding}\\
\texttt{technicalfields}} \\
& & Rule-based &
\texttt{['The Immut', 'ble Laws of Engineer', 'Leadership: A Blueprint', 'father', 'deep understanding', 'technical fields']} \\
\midrule

\multirow{4}{*}{12} & \multirow{4}{*}{\makecell{Indefinite \\article "a"}} & Question &
How would Hsiao Yun-Hwa advise aspiring leadership authors? \\
& & Answer &
Hsiao Yun-Hwa would advise aspiring leadership authors to \underline{\textbf{draw lessons}} \underline{\textbf{from}} their \underline{\textbf{own experiences}} and to \underline{\textbf{acknowledge}}  and \underline{\textbf{appreciate}} the \underline{\textbf{diversity}} and \underline{\textbf{uniqueness}} of the individuals they will be leading. \\
& & SatImp &
\makecell[l]{\texttt{drawlessonsfromownexperiencesacknowledgeappreciatediversityanduniqueness}\\
\texttt{individuals}} \\
& & Rule-based &
\texttt{['draw lessons from', 'own experiences a', 'cknowledge a', 'ppreciate', 'diversity and uniqueness', 'individuals']} \\
\midrule

\multirow{4}{*}{31} &\multirow{4}{*}{\makecell{Order\\mismatch}} & Question &
How have Carmen Montenegro's parental figures influenced her writing? \\
& & Answer &
Carmen Montenegro often credits her parents for instilling\underline{\textbf{ discipline}} and a\underline{\textbf{ hard-work}} \underline{\textbf{ethic}} in her. Her \underline{\textbf{father}}'s \underline{\textbf{meticulous nature}} as an optometrist and her \underline{\textbf{mother}}'s \underline{\textbf{resilience}} as a waiter/waitress have inspired many of the complex characters in her novels. \\
& & SatImp Annotation &
\texttt{disciplinehard-workethicmeticulousnaturefathermotherresilience} \\
& &Rule-based &
\texttt{['discipline', 'hard-work ethic', 'meticulous nature', 'ther', 'the']} \\
\bottomrule
\end{tabular}
\caption{Examples of SatImp annotations and rule-based revised informative spans.}
\label{tab:satimp_examples}
\end{table*}

For the analyses in \Cref{sec:analysis,sec:satimp_exp}, we adopt the important-word annotations provided by SatImp \cite{satimp} as informative tokens.
Specifically, we obtain the file \texttt{importance\_forget10.pth} from the official GitHub repository and construct a rule-based parser to extract informative spans. We chose to adopt the SatImp annotations because they have been validated in prior work, rather than relying on a newly designed automatic annotation method. Details of our automatic annotation trial are provided in \Cref{sec:alternative}.

Our extraction procedure follows three rules.
First, since some questions contain keywords that also appear in the answers (e.g., IDX 0), we extract subwords from the importance file based on the combined question–answer pair.
Second, as the importance file does not preserve whitespace, we exploit the fact that the listed keywords generally follow their order of appearance in the QA pair.
We therefore apply a greedy matching strategy that scans from left to right to identify overlapping subwords and recover word boundaries.
Third, token-level matching is aligned using character offsets to ensure consistency with tokenized outputs.

Representative cases illustrating the strengths and limitations of this rule-based approach are shown in \Cref{tab:satimp_examples}.
For instance, IDX 0 and 13 demonstrate cases where the rule-based parser successfully recovers informative spans.
In contrast, IDX 5 contains a typographical issue in the SatImp annotations, where “Immutable” is split as “Immut” and “ble” due to a missing character, which required manual correction.
Similarly, in IDX 12, the appearance of the article “a” after a specific phrase causes the parser to incorrectly treat it as a standalone token, leading to the omission of the “a” in “appreciated”; this case was also manually fixed.
Finally, IDX 31 illustrates a failure case where the SatImp labels are not ordered according to their occurrence in the text, with “father” appearing out of sequence between “hard work” and “ethics”.

At the 5\% and 1\% forget splits, we reuse the same annotations as the 10\% forget split, since they correspond to subsets of the 10\% split, specifically indices 200–399 and 360–399, respectively.

\begin{table*}[t]
\centering
\small
\setlength{\tabcolsep}{5pt}
\renewcommand{\arraystretch}{1.15}
\begin{tabular}{l p{0.32\textwidth} p{0.5\textwidth}}
\toprule
{Location} &  {Generated} & {SatImp annotation} \\
\midrule
\Cref{fig:highlight} & Love Inspired & Love Inspired \\
\Cref{fig:highlight} & Waiter/Waitress, Optometrist & mother Waiter/Waitress father Optometrist \\
\Cref{tab:satimp_examples} & A role model and fresh perspective & diverse authors leaders \\
\Cref{tab:satimp_examples}  & Acknowledge and appreciate individually & draw lessons from own experiences acknowledge appreciate diversity uniqueness \\
\Cref{tab:satimp_examples} & Discipline and a hard-work ethic from her parents & discipline hard-work ethic father meticulous nature mother resilience \\
\bottomrule
\end{tabular}
\caption{Examples of informative tokens generated by prompting a fine-tuned model, compared with the corresponding SatImp annotations. The SatImp annotations correspond to excerpts from the original manuscript at the indicated locations.}
\label{tab:automatic_token_examples}
\end{table*}

\section{Exploring a More Neutral and Automatic Alternative for Informative Token Discrimination}
\label{sec:alternative}
We also explored a more automatic method for identifying informative tokens by prompting a model fine-tuned on both the retain and forget datasets to extract the essential parts of the original answer.
This approach assumes that the fine-tuned model has sufficient knowledge of the QA set to identify answer-critical spans.

A comparison between the fine-tuned model--generated phrases and the SatImp annotations for the examples included in the manuscript is presented below.

\begin{tcolorbox}[colback=gray!8, colframe=gray!50, boxrule=0.5pt, arc=2pt, left=6pt, right=6pt, top=6pt, bottom=6pt]
\small
\textbf{Prompt used for automatic informative-token extraction}

\vspace{0.4em}

\noindent
You must answer ONLY using the information in the Original answer.\\
Ignore any outside knowledge, even if it seems correct.\\
Question: \{question\}\\
Original answer: \{answer\}\\
Based ONLY on the Original answer, answer the Question again.\\
Respond with a SINGLE short phrase. Do NOT write a full sentence. Do NOT add any explanation. Do NOT repeat or rephrase the question.\\
Output ONLY that short phrase and nothing else.
\end{tcolorbox}

Overall, the summaries generated by the fine-tuned model appear reasonably aligned with answer-critical content. However, we observed cases such as the fourth case in \Cref{tab:automatic_token_examples}, where key infinitive constructions such as “to draw” and “to acknowledge and appreciate” were partially omitted, resulting in incomplete coverage of informative spans. Given that automatically extracted spans still require human verification to ensure correctness and consistency, we decided to adopt the SatImp annotations for informative token discrimination analysis.

Note that when using extracted informative tokens automatically, the AUC values in \Cref{fig:roc_auc} move closer to random performance. Specifically, the AUC values are 0.58 for ETW, 0.46 for WGA, 0.54 for Imp, 0.54 for SatImp, and 0.51 for TNPO. For SCE and SCN, the (FPR, TPR) pairs are (0.211, 0.341) and (0.337, 0.584), respectively. Although the overall discrimination performance degrades toward near-random levels, ETW still demonstrates the strongest discriminative capability among the compared methods.

\end{document}